\documentclass{article} 
\usepackage{iclr2027_conference,times}

\usepackage{amsmath,amsfonts,bm}

\def\eqref#1{equation~\ref{#1}}

\def\1{\bm{1}}

\DeclareMathAlphabet{\mathsfit}{\encodingdefault}{\sfdefault}{m}{sl}
\SetMathAlphabet{\mathsfit}{bold}{\encodingdefault}{\sfdefault}{bx}{n}

\usepackage{hyperref}
\usepackage{url}
\usepackage{booktabs}
\usepackage{arydshln}
\usepackage{multirow}
\usepackage{graphicx}
\usepackage{hyperref}
\usepackage{url}
\usepackage{enumitem}
\usepackage{graphicx}
\usepackage{wrapfig}
\usepackage{algorithm}
\usepackage{algpseudocode}

\title{Generalist Representation, Specialist Detection: TS-Router for Time-Series Anomaly Detection}

\author{%
	Tian Lan, Yifei Gao, Yimeng Lu \& Xuming An \\
	Department of Industrial Engineering \\
	Tsinghua University \\
	\texttt{gao-yf@mail.tsinghua.edu.cn} \\
	\texttt{\{lant23,luym25,axm24\}@mails.tsinghua.edu.cn}
	\AND
	Meng Wang, Yue Pan \& Wenjun He \\
	Huawei \\
	\texttt{\{wangmeng71,panyue33,hewenjun8\}@huawei.com}
	\AND
	Chenghao Liu\thanks{%
		Corresponding authors: Chenghao Liu and Chen Zhang.
		Chenghao Liu's contribution was completed prior to joining Datadog.
	} \\
	Datadog AI Research \\
	Paris, France \\
	\texttt{twinsken@gmail.com}
	\And
	Chen Zhang\footnotemark[1] \\
	Department of Industrial Engineering \\
	Tsinghua University \\
	\texttt{zhangchen01@tsinghua.edu.cn}
}
\iclrfinalcopy

\begin{document}

	\maketitle
	\lhead{}
	\begin{abstract}
		Time-series anomaly detection (TSAD) is difficult to generalize across datasets because heterogeneous temporal dynamics imply different notions of normality and favor different detection criteria. While time-series foundation models provide transferable representations, coupling them with a fixed anomaly-scoring mechanism can overlook this variation. This motivates a different perspective on foundation-model-based TSAD: using foundation models to coordinate specialized anomaly criteria rather than directly imposing a universal one. Based on this view, we propose \textbf{TS-Router}, a generalist-representation, specialist-detection framework that estimates the relative competence of heterogeneous anomaly detectors from pretrained temporal representations and selects suitable specialists for each target series. To avoid relying on specialist-performance labels from real tasks, we derive soft competence supervision from specialists' relative performance on labeled simulated tasks. At deployment, routing requires no target anomaly labels, and only the selected specialists are fitted unsupervisedly on the target series. We bound Top-\(k\) set-competence regret under representation coverage and conditional competence stability. Across 16 real-world benchmarks and four complementary evaluation metrics, TS-Router achieves the best overall average rank. Controlled ablations with multiple frozen TSFM encoders further support the use of pretrained representations for competence estimation and adaptive specialist selection. The code is available at \url{https://anonymous.4open.science/r/TS-Router-D8FF}. 
	\end{abstract}
	
	\section{Introduction}
	
	Time-series anomaly detection (TSAD) is a fundamental task in sequential data analysis, with broad applications in industrial monitoring \citep{nizam2022real,chen2022deep}, system operations \citep{audibert2020usad,guo2024logformer}, health care \citep{su2019robust,xu2021anomaly}, and transportation and energy systems \citep{deng2021graph}.
	TSAD aims to identify observations or subsequences that are inconsistent with the normal behavior of the underlying temporal process, typically in settings where anomalies are rare and reliable anomaly annotations are unavailable.
	Accordingly, reliable TSAD requires two coupled capabilities: modeling the contextual regularities that characterize normal temporal behavior, and assessing whether deviations from these regularities indicate anomalies or merely reflect normal variability \citep{mueller2025open,wang2025survey}.
	
	Conventional TSAD methods learn normal behavior separately for each target series and score deviations from it. For example, PCA-based detectors fit a low-dimensional subspace and flag observations with large reconstruction errors \citep{yairi2001fault}. This target-specific design leads to a one-dataset-one-model paradigm, requiring a detector to be selected, tuned, and fitted for every target. Yet detector suitability varies across series, since different temporal dynamics favor different anomaly criteria. Detector selection therefore depends on domain expertise and repeated trial and error. Automated selection methods alleviate this burden by choosing among candidate detectors based on target characteristics \citep{zhao2021automatic,goswami2022unsupervised,sylligardos2023choose,schmidl2024autotsad}. However, they typically rely on handcrafted descriptors, limited historical performance records, or deployment-time candidate evaluation. This exposes a complementary challenge: while detector suitability is target-dependent, inferring such suitability on an unseen series requires a representation of temporal structure that transfers across datasets.
	
	Time-series foundation models (TSFMs) offer a natural way to address this representational challenge by learning transferable temporal representations from large-scale corpora that can be readily applied to new target series. This capability has motivated growing interest in applying models such as MOMENT and TimesFM to TSAD \citep{goswami2024moment,das2024decoder}. Most existing TSFM-based approaches, however, couple these transferable representations with forecasting or reconstruction objectives and use the resulting error as a fixed anomaly criterion. A recent systematic evaluation shows that such direct adaptations may perform no better than simple local-statistics baselines across diverse time series \citep{zhu2026foundation}. Rather than negating the value of pretrained temporal representations, these findings suggest that solving the representation problem does not necessarily solve the anomaly-decision problem, raising a more fundamental question: \textbf{beyond directly defining an anomaly criterion, what role should foundation models play in TSAD?}
	
	\begin{wrapfigure}{r}{0.35\textwidth}
		\centering
        \vspace{-10pt}
		\includegraphics[width=\linewidth]{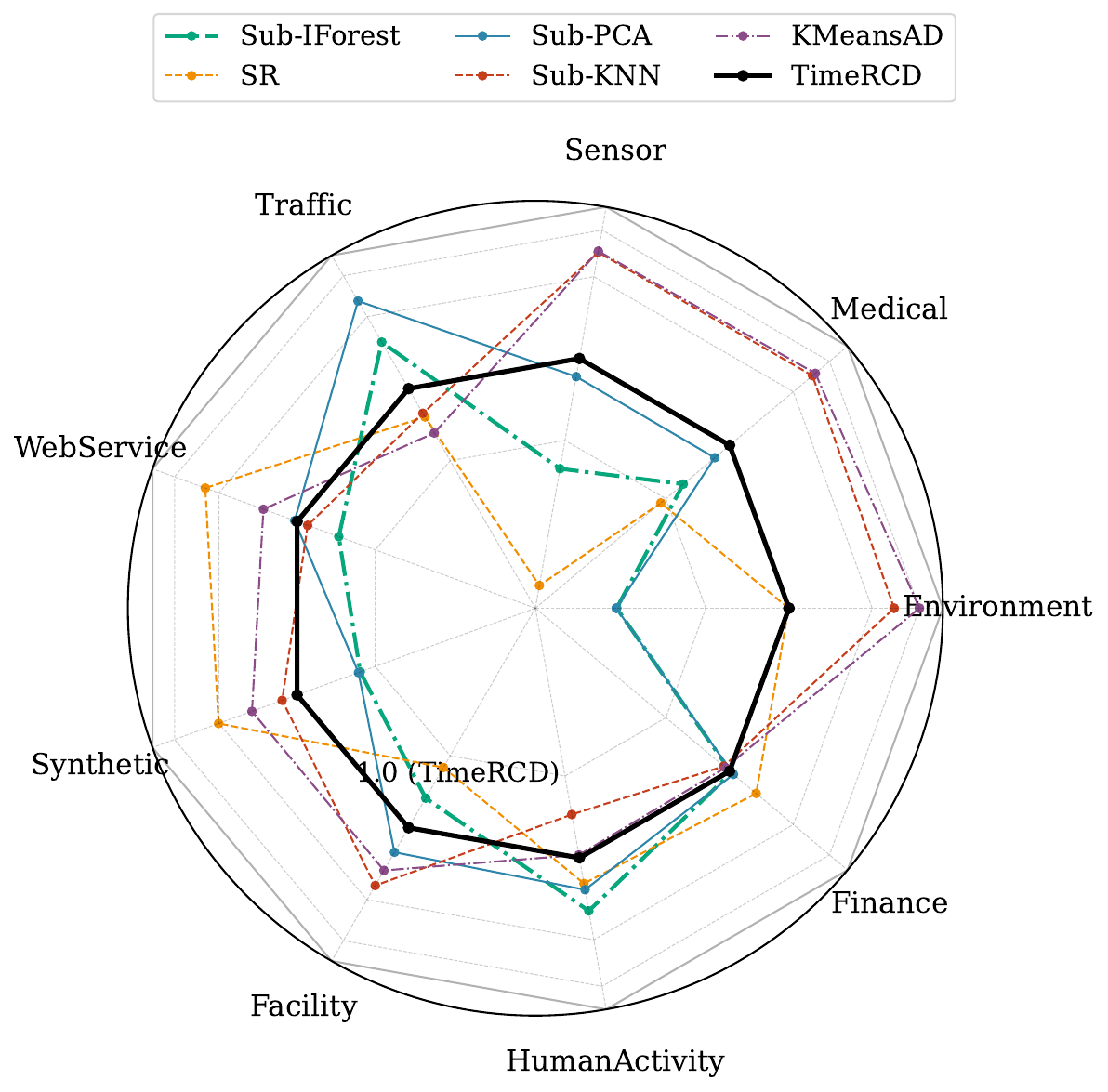}
        \vspace{-15pt}
		\caption{Domain-wise VUS-PR of representative specialists relative to TimeRCD, normalized by TimeRCD performance.}
		\label{fig:specialist_complementarity}
        \vspace{-10pt}
	\end{wrapfigure}
	
	This question reflects a fundamental difference between forecasting and anomaly detection. Autoregressive forecasting admits a shared objective, such as \(\mathbf{x}_{1:t} \mapsto \mathbf{x}_{t+1:t+H}\), where each observed continuation provides supervision, enabling pretraining on abundant unlabeled series. TSAD, in contrast, must determine which departures from normal behavior constitute anomalies, despite scarce annotations and anomaly criteria whose suitability varies across targets. The same pattern may be normal for one series but anomalous for another, and different detectors may capture different forms of deviation. As shown in Figure~\ref{fig:specialist_complementarity}, the relative strengths of representative specialists and the anomaly-specific foundation model TimeRCD \citep{lan2025towards} vary substantially across domains. Together, these observations motivate conditional specialization and suggest a separation between what can be transferred and what should remain adaptive: temporal representations may generalize across series, whereas the choice of anomaly criterion can benefit from target-dependent adaptation.
	
	Motivated by this separation, we develop \textbf{TS-Router}. Given an unlabeled target series, a frozen foundation encoder represents its temporal context, and a learned router predicts the relative competence of candidate specialists. Only the selected detectors are then fitted unsupervisedly on the target series, and their normalized anomaly scores are combined for detection. This route-before-fit procedure avoids fitting every candidate at deployment and requires no target anomaly labels. Our main contributions are:
	
	\begin{itemize}[leftmargin=1em]
		\item We introduce a generalist-representation, specialist-detection perspective for TSAD, separating transferable temporal representation from target-dependent anomaly decision and positioning foundation models as coordinators of specialized detection criteria through competence routing.
		
		\item We propose \textbf{TS-Router}, which uses transferable temporal representations to estimate heterogeneous detector competence and adaptively select and combine suitable specialists. Without large-scale real anomaly annotations, we derive scalable soft routing supervision from specialists' relative performance on labeled simulated series, and establish a Top-\(k\) set-competence regret bound under explicit transfer assumptions.
		
		\item Extensive experiments on diverse TSAD benchmarks show that TS-Router achieves the best overall average rank among the evaluated baselines. Controlled routing and representation ablations demonstrate the benefits of adaptive specialist selection and show that the framework is effective with multiple frozen TSFM encoders.
	\end{itemize}
	
	\section{Related Work}
	
	\textbf{Target-Specific Time-Series Anomaly Detection.}
	Most unsupervised TSAD methods are fitted separately to each target series and identify anomalies as deviations from target-specific normality. Deep approaches commonly use reconstruction objectives, as in USAD, OmniAnomaly, TranAD, and TFMAE \citep{audibert2020usad,su2019robust,tuli2022tranad,fang2024temporal}, or learn discriminative representations through self-supervised and contrastive objectives \citep{yang2023dcdetector,xu2021anomaly,shen2020timeseries,darban2025dacad,lan2025cicada}. Classical detectors based on statistical, distance, density, subspace, isolation, and one-class criteria also remain competitive \citep{goldstein2012histogram,li2007unifying,ramaswamy2000efficient,breunig2000lof,yairi2001fault,scholkopf1999support,liu2008isolation,ren2019time}. These methods embody different inductive biases for modeling normality and scoring deviations, whose suitability can vary substantially across target series. Consequently, deployment still requires target-specific detector selection and adaptation, limiting scalable generalization across heterogeneous series.
	
	\textbf{Time-Series Foundation Models for Anomaly Detection.}
	Time-series foundation models learn transferable temporal representations through large-scale pretraining. General-purpose models such as MOMENT and UniTS support multiple downstream tasks, while TimesFM, Chronos, and Time-MoE primarily transfer forecasting capabilities across datasets \citep{goswami2024moment,gao2024units,das2024decoder,ansari2024chronos,shi2024time}. Pretrained models have also been adapted to anomaly detection, including DADA and TSPulse \citep{shentu2024towards,ekambaram2025tspulse}. More recently, TimeRCD develops an anomaly-specific foundation model pretrained with labeled synthetic data and achieves strong zero-shot performance \citep{lan2025towards}. Despite their transferable representations, these approaches generally couple them with a unified detector or fixed anomaly-scoring mechanism across heterogeneous targets. Direct use of forecasting or reconstruction errors may also perform comparably to simple local-statistics baselines, since anomalies are not consistently harder to forecast or reconstruct \citep{zhu2026foundation}. These observations leave open whether pretrained temporal representations can instead support target-dependent adaptation of anomaly criteria across heterogeneous series.
	
	\textbf{Automated Detector Selection.}
	Automated detector-selection methods address detector heterogeneity by identifying suitable candidates for unlabeled targets. MetaOD transfers historical detector performance through meta-learning and handcrafted descriptors \citep{zhao2021automatic}, while Unsupervised Model Selection and Choose Wisely assess detector suitability using surrogate criteria, candidate outputs, or series characteristics \citep{sylligardos2023choose,goswami2022unsupervised}. AutoTSAD further automates detector configuration, selection, and ensembling for individual targets \citep{schmidl2024autotsad}. Meanwhile, pseudo-anomaly and anomaly-injection methods use synthetic samples primarily to train anomaly detectors directly \citep{obata2025robust,jeong2023anomalybert,darban2025carla,shentu2024towards}. Existing selection methods, however, remain tied to handcrafted descriptors, historical task collections, surrogate evaluations, or deployment-time candidate execution, limiting transferable and scalable competence estimation on unseen series.
	
	\section{Problem Formulation}
	\label{sec:problem}
	
	\textbf{TSAD with Heterogeneous Specialists.}
	Let \(\mathcal T=(\mathbf X,\mathbf y)\) denote a TSAD task, where \(\mathbf X=(x_1,\ldots,x_L)\in\mathbb R^L\) is the observed series and \(\mathbf y\in\{0,1\}^L\) contains its anomaly labels. We formulate the selection problem for univariate series; multivariate targets use variable-wise routing followed by score aggregation. Target labels are unavailable at deployment and are used only to define or evaluate competence. Consider a pool \(\mathcal M=\{M_1,\ldots,M_K\}\) of heterogeneous unsupervised detectors. Each \(M_j\) is a \emph{specialist} with a particular inductive bias for modeling normality and scoring deviations, rather than a detector assigned to a predefined anomaly type. Under a fixed label-free fitting and scoring protocol, it produces scores \(\mathbf a_j(\mathbf X)\) with competence \(c_j(\mathcal T)=\Phi(\mathbf a_j(\mathbf X),\mathbf y)\). The utility \(\Phi\) is evaluated on the protocol's evaluation interval and expressed on the \([0,1]\) scale. The vector \(\mathbf c(\mathcal T)=[c_1(\mathcal T),\ldots,c_K(\mathcal T)]^\top\) records specialist suitability.
	
	\textbf{Specialist Selection with a Fixed Budget.}
	For a budget \(1\le k\le K\), let \(\mathcal I_k=\{S\subseteq[K]:|S|=k\}\), where \([K]=\{1,\ldots,K\}\). We assess a selected set through its mean individual competence. For \(\mathbf v\in\mathbb R^K\), write
	\begin{equation}
		V_k(S,\mathbf v)=\frac1k\sum_{j\in S}v_j,
		\qquad
		\mathsf T_k(\mathbf v)=\max_{S\in\mathcal I_k}V_k(S,\mathbf v).
		\label{eq:topk_set_utility}
	\end{equation}
	The maximizing set is \(\operatorname{TopK}(\mathbf v,k)\), with a fixed tie-breaking rule. The best globally fixed set and hindsight task-wise selection attain
	\begin{equation}
		U_{\mathrm{fixed}}^{(k)}=\max_{S\in\mathcal I_k}\mathbb E[V_k(S,\mathbf c(\mathcal T))],
		\qquad
		U_{\mathrm{oracle}}^{(k)}=\mathbb E[\mathsf T_k(\mathbf c(\mathcal T))].
		\label{eq:topk_reference_utilities}
	\end{equation}
	These utilities compare sets of the same size before score fusion; they are not the detection utility of an averaged score sequence. When \(k=1\), they reduce to individual-specialist selection.
	
	\textbf{Representation-Conditioned Competence.}
	Although \(\mathbf c(\mathcal T)\) is unobserved at deployment, temporal information can guide selection. Let \(Z=E(\mathbf X)\) and \(\boldsymbol\mu(z)=\mathbb E[\mathbf c(\mathcal T)\mid Z=z]\). The best budget-\(k\) selection based on \(Z\) attains \(U_{\mathrm{cond}}^{(k)}(Z)=\mathbb E_Z[\mathsf T_k(\boldsymbol\mu(Z))]\), and
	\begin{equation}
		U_{\mathrm{fixed}}^{(k)}\le U_{\mathrm{cond}}^{(k)}(Z)\le U_{\mathrm{oracle}}^{(k)}.
		\label{eq:utility_hierarchy}
	\end{equation}
	Thus, observable context can improve selection over a fixed set, while \(U_{\mathrm{oracle}}^{(k)}-U_{\mathrm{cond}}^{(k)}(Z)\) measures the information gap to hindsight selection. If \(Z_0=h(Z_1)\), then \(U_{\mathrm{cond}}^{(k)}(Z_1)\ge U_{\mathrm{cond}}^{(k)}(Z_0)\). Transferable representations therefore need not directly define a universal anomaly criterion: they can instead inform which specialist criteria to deploy.
	
	\textbf{Competence Routing.}
	A router \(g_\theta\) maps \(z=E(\mathbf X)\) to a predicted competence distribution \(\hat{\mathbf q}_\theta(z)\in\Delta^{K-1}\). Its Top-\(k\) index set is \(\widehat S_\theta^{(k)}(z)=\operatorname{TopK}(\hat{\mathbf q}_\theta(z),k)\), corresponding to the specialist set \(\mathcal S_\theta(\mathbf X)=\{M_j:j\in\widehat S_\theta^{(k)}(E(\mathbf X))\}\). Routing precedes detector fitting. Only these specialists are fitted without target labels, and their normalized scores are combined for detection. This separates representation-conditioned specialist selection from the subsequent use of their detection outputs.
	
	\section{Method}
	\label{sec:method}
	
	\begin{figure}[t]
		\centering
		\includegraphics[width=\linewidth]{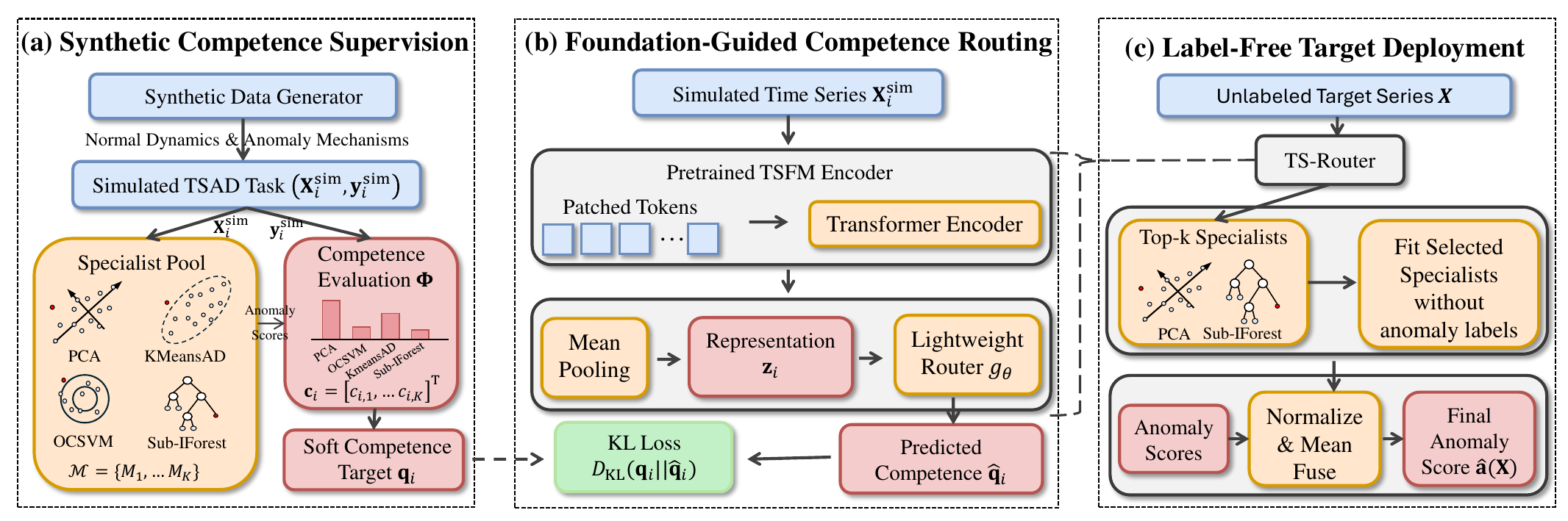}
		\vspace{-17pt}
		\caption{Overview of the TS-Router framework.}
		\vspace{-12pt}
		\label{fig:framework}
	\end{figure}
	
	As illustrated in Figure~\ref{fig:framework}, TS-Router learns specialist selection from temporal representations and synthetic soft competence supervision. Training follows two stages: a time-series foundation model first learns a temporal representation under its native objective; the encoder is then frozen, and a lightweight router is trained to predict specialist competence from the fixed representation. At deployment, routing precedes detector fitting, after which only the selected specialists are fitted without anomaly labels and combined for detection. This separates general temporal representation learning from target-adaptive anomaly decision.
	
	\subsection{Synthetic Competence Supervision}
	\label{sec:synthetic_competence}
	
	Real-world TSAD data provide insufficient supervision for learning how specialist competence varies across temporal contexts, as anomaly labels are scarce and anomalies are rare. Following the labeled simulation setting in RCD \citep{lan2025towards}, we construct a simulated task collection \(\mathcal{D}_{\mathrm{sim}}=\{(\mathbf{X}_i^{\mathrm{sim}},\mathbf{y}_i^{\mathrm{sim}})\}_{i=1}^{N}\) covering diverse temporal structures and anomaly manifestations. Conceptually, the simulator induces a task distribution \(P_s(\mathcal{T})\) over competence-relevant variations in temporal structures and anomaly mechanisms, rather than attempting to reproduce the real data distribution itself. This construction varies both normal temporal dynamics and anomaly mechanisms, since either can change the competence ordering among specialists. Each simulated task therefore serves as a \emph{competence probe} for the specialist pool. The specialist pool is fixed across competence construction and target deployment and intentionally consists of lightweight detectors with heterogeneous inductive biases. This design makes routing primarily reflect which notion of normality and deviation is appropriate, rather than differences in model capacity. The complete pool and its corresponding inductive biases are detailed in Appendix~\ref{app:specialists}.
	
For each simulated task, every specialist $M_j$ is evaluated under
the Syn-RCD label-free fitting and scoring protocol and produces
anomaly scores $a_{i,j}$. Since the simulated labels are known, its competence is evaluated as \(c_{i,j}^{\mathrm{sim}}=\Phi(\mathbf{a}_{i,j},\mathbf{y}_i^{\mathrm{sim}})\), where \(\Phi\) is instantiated as VUS-PR and expressed on the \([0,1]\) scale for competence construction. Thus, anomaly labels are used to evaluate specialist competence rather than to fit the specialists. The resulting profile \(\mathbf{c}_i^{\mathrm{sim}}=[c_{i,1}^{\mathrm{sim}},\ldots,c_{i,K}^{\mathrm{sim}}]^\top\) characterizes the relative suitability of different specialist biases for task \(i\). Instead of reducing this profile to a single best-specialist label, we transform it into a soft competence target \(\mathbf{q}_i\in\Delta^{K-1}\):
	\begin{equation}
		q_{i,j}
		=
		\frac{\exp(c_{i,j}^{\mathrm{sim}}/\tau)}
		{\sum_{\ell=1}^{K}\exp(c_{i,\ell}^{\mathrm{sim}}/\tau)},
		\qquad j=1,\ldots,K,
		\label{eq:soft_competence}
	\end{equation}
	where \(\tau>0\) controls the sharpness of the target distribution. Soft targets preserve each task's competence ordering and encode pairwise competence gaps through their log-ratios, including cases in which several specialists exhibit comparable competence. The resulting routing dataset \(\mathcal{D}_{\mathrm{route}}=\{(\mathbf{X}_i^{\mathrm{sim}},\mathbf{q}_i)\}_{i=1}^{N}\) provides scalable supervision for learning how specialist suitability varies with observable temporal context.
	
\subsection{Foundation-Guided Competence Routing}
\label{sec:competence_routing}

We combine a pretrained TSFM encoder with a trainable competence router.
The encoder \(E_{\mathrm{TSFM}}\) is obtained through pretraining under its
native objective, independently of the specialist competence supervision
used for routing.

During routing training, \(E_{\mathrm{TSFM}}\) is frozen and used only as a
feature extractor. Given a simulated series
\(\mathbf{X}_i^{\mathrm{sim}}\), the final encoder layer produces
contextualized token embeddings
\(\mathbf{H}_i=[\mathbf{h}_{i,1},\ldots,\mathbf{h}_{i,m}]^\top
\in\mathbb{R}^{m\times D}\), where \(m\) is the number of temporal tokens
and \(D\) is the representation dimension. Since competence routing is
performed at the series level, we mean-pool all final-layer token
embeddings to obtain
\[
\mathbf{z}_i
=
\frac{1}{m}\sum_{j=1}^{m}\mathbf{h}_{i,j}
\in\mathbb{R}^{D}.
\]
The foundation model therefore provides a fixed temporal representation
shared across specialists for inferring their suitability, rather than
directly producing anomaly scores.

A lightweight router \(g_\theta\), implemented as an MLP, maps
\(\mathbf{z}_i\) to \(K\) specialist logits and predicts the competence
distribution
\begin{equation}
	\hat{\mathbf{q}}_i
	=
	\operatorname{Softmax}\!\left(g_\theta(\mathbf{z}_i)\right)
	\in\Delta^{K-1},
	\label{eq:router_prediction}
\end{equation}
where \(\hat{q}_{i,j}\) represents the predicted relative competence of
specialist \(M_j\). Predicting the full competence distribution preserves
relative suitability across multiple specialists and aligns naturally with
the soft targets defined in Section~\ref{sec:synthetic_competence}.

With \(E_{\mathrm{TSFM}}\) fixed, only the router parameters \(\theta\) are
optimized by matching \(\hat{\mathbf{q}}_i\) to \(\mathbf{q}_i\) using the
Kullback--Leibler divergence:
\begin{equation}
	\mathcal{L}_{\mathrm{route}}
	=
	\frac{1}{N}
	\sum_{i=1}^{N}
	D_{\mathrm{KL}}
	\left(
	\mathbf{q}_i
	\,\Vert\,
	\hat{\mathbf{q}}_i
	\right).
	\label{eq:routing_loss}
\end{equation}
The resulting router predicts soft competence targets from fixed temporal
representations.

\subsection{Offline Target Routing and Detection}
\label{sec:deployment}

We study offline anomaly detection on an observed target series
\(\mathbf{X}=(\mathbf{X}^{\mathrm{tr}},\mathbf{X}^{\mathrm{te}})\), where the
within-series split designates a specialist-fitting prefix and an
evaluation suffix. Both portions are available as unlabeled observations
when routing is performed. The prefix specifies the observations used for
specialist fitting; it does not delimit the input available to the router.
Routing therefore conditions on the observed evaluation suffix as well as
the prefix, while target anomaly labels are unavailable to routing,
specialist fitting, and score generation.

The encoder is pretrained under its native objective, and the router is
trained on \(\mathcal{D}_{\mathrm{route}}\). Both models are frozen before
target deployment and process the complete unlabeled observed series only
for forward inference; no target labels or encoder/router parameter
updates are used. The router predicts the competence distribution
\(\hat{\mathbf{q}}(\mathbf{X})\), from which we select
\[
\mathcal{S}_{\theta}(\mathbf{X})
=
\left\{
M_j:
j\in\operatorname{TopK}
\bigl(\hat{\mathbf{q}}(\mathbf{X}),k\bigr)
\right\}.
\]

Each selected specialist \(M_j\in\mathcal{S}_{\theta}(\mathbf{X})\) is fitted
to \(\mathbf{X}^{\mathrm{tr}}\) without anomaly labels and produces a score
sequence \(\mathbf{a}_j(\mathbf{X}^{\mathrm{te}})\) on the evaluation
suffix. We standardize each sequence, average the selected outputs, and
min--max rescale the fused scores:
\begin{equation}
	\hat{\mathbf{a}}^{\mathrm{te}}
	=
	\mathcal{M}\!\left(
	\frac{1}{k}
	\sum_{M_j\in\mathcal{S}_{\theta}(\mathbf{X})}
	\mathcal{N}\!\left(
	\mathbf{a}_j(\mathbf{X}^{\mathrm{te}})
	\right)
	\right).
	\label{eq:topk_fusion}
\end{equation}

Here, \(\mathcal{N}\) and \(\mathcal{M}\) use statistics of the
corresponding score sequences on the evaluation suffix, as specified in
Appendix~\ref{app:algorithm}. These operations produce continuous anomaly
scores without target labels. For multivariate targets, the same
observation protocol and routing--fitting--scoring procedure are applied
independently to each variable, followed by the cross-variable aggregation
described in that appendix. Algorithm~\ref{alg:tsrouter} summarizes the
complete procedure.
	\subsection{Theoretical Analysis}
	\label{sec:theory}
	
We analyze Top-\(k\) selection at the deployment budget using the set
competence in Eq.~\ref{eq:topk_set_utility}. Let \(P_s\) and \(P_r\)
denote simulated and real task distributions, with
\(Z=E(\mathbf X)\) denoting the fixed, mean-pooled TSFM representation.
Throughout, \(K\ge2\), \(1\le k\le K\), \(\tau>0\), and
\(\mathbf c(\mathcal T)\in[0,1]^K\). Write
\(\mathbf q(\mathcal T)=\operatorname{Softmax}(\mathbf c(\mathcal T)/\tau)\)
and \(\mathbf p_\theta(z)=\hat{\mathbf q}_\theta(z)\). The router is
measurable and strictly positive, and selection uses a fixed measurable
tie-breaking rule. Logarithms are natural, and
\(\operatorname{osc}(\mathbf v)=\max_jv_j-\min_jv_j\).
	
	\textbf{Lemma 1 (Top-\(k\) Competence Regret from KL).}
	Set \(m_k=\min\{k,K-k\}\), \(\alpha_k=m_k/k\), and \(D_{\tau,K,k}=(k-1)e^{1/\tau}+K-k+1\). For \(\operatorname{osc}(\mathbf v)\le1\), \(\mathbf q=\operatorname{Softmax}(\mathbf v/\tau)\), and any strictly positive probability vector \(\mathbf p\),
	\begin{equation}
		0\le\mathsf T_k(\mathbf v)-V_k(\operatorname{TopK}(\mathbf p,k),\mathbf v)
		\le\min\!\left\{\alpha_k,\;C_{\tau,K,k}\sqrt{2D_{\mathrm{KL}}(\mathbf q\Vert\mathbf p)}\right\},
		\label{eq:kl_regret}
	\end{equation}
	where
	\begin{equation}
		C_{\tau,K,k}=\min\!\left\{
		\frac{D_{\tau,K,k}}{k(1-e^{-1/\tau})},\;
		\frac{\sqrt{m_kD_{\tau,K,k}}}{k(1-e^{-1/(2\tau)})}
		\right\}.
		\label{eq:topk_regret_constant}
	\end{equation}
	The proof pairs missed and incorrectly selected specialists and uses the lower bound \(q_{(k)}\ge D_{\tau,K,k}^{-1}\) on the \(k\)-th largest target probability.
	
	Define \(\boldsymbol\mu_d(z)=\mathbb E_{P_d}[\mathbf c(\mathcal T)\mid Z=z]\), \(d\in\{s,r\}\), and \(\mathbf m_s(z)=\mathbb E_{P_s}[\mathbf q(\mathcal T)\mid Z=z]\). Let \(\mathcal L_s^*\) be the infimum of the population routing loss over all measurable simplex-valued predictors. The loss and its excess satisfy
	\begin{align}
		\mathcal L_s(\theta)
		&=\mathbb E_{P_s}D_{\mathrm{KL}}(\mathbf q(\mathcal T)\Vert\mathbf p_\theta(Z)),
		\label{eq:population_routing_loss}\\
		\mathcal E_s(\theta):=\mathcal L_s(\theta)-\mathcal L_s^*
		&=\mathbb E_{P_s^Z}D_{\mathrm{KL}}(\mathbf m_s(Z)\Vert\mathbf p_\theta(Z)).
		\label{eq:routing_excess_risk}
	\end{align}
	The minimizer is \(\mathbf m_s\), which need not equal \(\operatorname{Softmax}(\boldsymbol\mu_s/\tau)\). We account for this difference through
	\begin{equation}
		b_{\tau,s}(z)=\operatorname{osc}(\boldsymbol\mu_s(z)-\tau\log\mathbf m_s(z)),
		\qquad
		\beta_{\tau,r}=\mathbb E_{P_r^Z}[b_{\tau,s}(Z)].
		\label{eq:conditional_softmax_distortion}
	\end{equation}
	Assume \(P_r^Z\ll P_s^Z\), with \(dP_r^Z/dP_s^Z\le\rho<\infty\), and \(\|\boldsymbol\mu_r(z)-\boldsymbol\mu_s(z)\|_\infty\le\delta\) for \(P_r^Z\)-almost every \(z\).
	
	The selected set has target competence \(U_{\mathrm{router},r}^{(k)}=\mathbb E_{P_r^Z}[V_k(\widehat S_\theta^{(k)}(Z),\boldsymbol\mu_r(Z))]\); a domain subscript specifies the distribution used to evaluate each utility.
	
	\textbf{Theorem 1 (Simulated-to-Real Top-\(k\) Routing Regret).}
	Under the stated assumptions, every router with finite \(\mathcal L_s(\theta)\) satisfies
	\begin{align}
		U_{\mathrm{cond},r}^{(k)}(Z)-U_{\mathrm{router},r}^{(k)}
		&\le\min\!\left\{\alpha_k,\;C_{\tau,K,k}\sqrt{2\rho\,\mathcal E_s(\theta)}
		+\alpha_k(\beta_{\tau,r}+2\delta)\right\},
		\label{eq:real_routing_regret}\\
		U_{\mathrm{cond},r}^{(k)}(Z)-U_{\mathrm{router},r}^{(k)}
		&\le\min\!\left\{\alpha_k,\;C_{\tau,K,k}\sqrt{2\rho\,\mathcal L_s(\theta)}+2\alpha_k\delta\right\}.
		\label{eq:total_routing_regret}
	\end{align}
	
	Let \(\epsilon_k(\theta)\) be the smaller right-hand side. Define the budget-specific representation gap and routing opportunity by
	\begin{equation}
		\Delta_{\mathrm{rep}}^{(k)}(Z)=U_{\mathrm{oracle},r}^{(k)}-U_{\mathrm{cond},r}^{(k)}(Z),
		\qquad
		\Gamma_{\mathrm{route}}^{(k)}(Z)=U_{\mathrm{cond},r}^{(k)}(Z)-U_{\mathrm{fixed},r}^{(k)}.
		\label{eq:representation_routing_gap}
	\end{equation}
	Then
	\begin{equation}
		\begin{aligned}
			U_{\mathrm{oracle},r}^{(k)}-U_{\mathrm{router},r}^{(k)}
			&\le\Delta_{\mathrm{rep}}^{(k)}(Z)+\epsilon_k(\theta),\\
			U_{\mathrm{router},r}^{(k)}-U_{\mathrm{fixed},r}^{(k)}
			&\ge\Gamma_{\mathrm{route}}^{(k)}(Z)-\epsilon_k(\theta).
		\end{aligned}
		\label{eq:utility_decomposition}
	\end{equation}
	The bound separates representation information, soft-target prediction, and competence transfer at a fixed budget. Proofs appear in Appendix~\ref{app:proof}. Appendix~\ref{app:selection_fusion} relates recovery of the competence-optimal set to the fused prediction when the selection boundary is separated.
	
	\section{Experiments}
	\label{sec:experiments}
	
	We evaluate TS-Router on 16 TSAD benchmarks to answer three questions:
	(1) \emph{Overall effectiveness:} can TS-Router generalize across heterogeneous targets and outperform foundation-model-based and target-fitted TSAD methods?
	(2) \emph{Routing effectiveness:} does competence-based routing improve over a globally fixed specialist, Best Top-3, and full-pool ensembling?
	(3) \emph{Representation effectiveness:} do TSFM representations provide stronger signals for specialist competence estimation than conventional features, and are the benefits consistent across pretrained encoders?
	Table~\ref{tab:main_results} addresses the first question across 16 benchmarks, while Table~\ref{tab:routing_effectiveness} examines the latter two on eleven univariate benchmarks to isolate routing and representation effects from variable-wise aggregation used for multivariate targets.
	Per-dataset results, implementation details, specialist complementarity, and
	representation analyses are provided in the appendix. Appendices~\ref{app:topk}--\ref{app:head_ablation} cover budget, pool-size,
	and routing-head sensitivity; transfer tests show that TS-Router leads
	classical selectors under synthetic-to-real and leave-one-dataset-out
	settings.
	\subsection{Experimental Setup}
	\label{sec:experimental_setup}
	
	\textbf{Datasets and Metrics.}
	We evaluate 16 TSAD benchmarks, including eleven genuinely univariate collections (IOPS, MGAB, NAB, NEK, Power, SED, Stock, TODS, UCR, WSD, and YAHOO) and five multivariate datasets (MSL, PSM, SMAP, SMD, and SWaT). Following TSAD evaluation practice, we report VUS-PR, Affiliation-F1, F1$_{\mathrm{T}}$, and Standard-F1, which capture complementary ranking-, event-, and point-level detection behavior. For routing analysis, NDCG@$k$ and Hit@$k$ measure agreement between the predicted specialist ranking and the VUS-PR-based ground-truth competence ranking. Detailed dataset statistics and preprocessing are provided in Appendix~\ref{app:datasets}. Threshold-dependent metrics use oracle thresholds selected from evaluation
	labels after score generation; these labels are never used for routing,
	specialist fitting, or anomaly-score generation.
	
	\textbf{Baselines and Deployment Protocols.}
	We compare TS-Router with direct zero-shot models and target-fitted unsupervised TSAD methods. Neither detector fitting nor deployment-time specialist selection uses target anomaly labels, although deployment protocols differ. Direct zero-shot models perform inference without target-specific detector fitting, whereas target-fitted methods and the specialists selected by TS-Router adapt unsupervisedly to each target series. Target anomaly labels are used only for evaluation, including hindsight reference strategies. Full baseline descriptions and implementation details are provided in Appendix~\ref{app:implementation}. For the controlled routing and representation comparisons, all applicable selectors and feature extractors receive the same unlabeled target observations. Specialist fitting data, evaluation intervals, and score-normalization and fusion rules are matched wherever applicable.
	
	\textbf{TS-Router Configuration.}
	TS-Router uses a fixed pool of eleven lightweight specialists with heterogeneous inductive biases. By default, we instantiate \(E_{\mathrm{TSFM}}\) with the frozen TimeRCD encoder, and construct routing supervision from specialists' VUS-PR competence on labeled Syn-RCD tasks. The same learned router is evaluated under all four detection metrics. At deployment, the router selects the Top-3 specialists before fitting; only these specialists are fitted unsupervisedly to the target series and their normalized scores are mean-fused. Representation ablations replace the default encoder with conventional feature representations or frozen Chronos and MOMENT encoders, training a router for each representation under the same competence supervision and deployment protocol. Unless otherwise stated, all experiments use the default configuration. Specialist definitions, hyperparameters, and encoder-specific implementation details are provided in the appendix.
	
	\begin{table}[t]
		\caption{
			Overall performance across 16 real-world TSAD benchmarks.
			Score denotes the mean performance across datasets, and Rank denotes
			the average dataset-wise rank among the 12 compared methods,
			with ties assigned average ranks.
			Overall Rank averages ranks across all four metrics.
			\#Top-1 and \#Top-2 count first-place and top-two finishes
			across the 64 dataset--metric combinations.
			Methods follow different deployment paradigms without using
			target anomaly labels for model fitting.
			Best results are in \textbf{bold} and second-best results are
			\underline{underlined}.
		}
		\label{tab:main_results}
		\centering
		\setlength{\tabcolsep}{2.5pt}
		\renewcommand{\arraystretch}{1.05}
		\resizebox{\textwidth}{!}{
			\begin{tabular}{@{}lccccccccccc@{}}
				\toprule
				\multirow{2}{*}{Method}
				& \multicolumn{2}{c}{VUS-PR}
				& \multicolumn{2}{c}{Aff.-F1}
				& \multicolumn{2}{c}{F1$_{\mathrm{T}}$}
				& \multicolumn{2}{c}{Std.-F1}
				& \multirow{2}{*}{\shortstack{Overall\\Rank}}
				& \multirow{2}{*}{\#Top-1}
				& \multirow{2}{*}{\#Top-2} \\
				\cmidrule(lr){2-3}
				\cmidrule(lr){4-5}
				\cmidrule(lr){6-7}
				\cmidrule(lr){8-9}
				& Score$\uparrow$ & Rank$\downarrow$
				& Score$\uparrow$ & Rank$\downarrow$
				& Score$\uparrow$ & Rank$\downarrow$
				& Score$\uparrow$ & Rank$\downarrow$
				& & & \\
				\midrule
				
				\multicolumn{12}{c}{\textit{Foundation-guided specialist routing}} \\
				\midrule
				\textbf{TS-Router}
				& \textbf{49.59} & \textbf{2.44}
				& \textbf{86.87} & \textbf{2.44}
				& \textbf{47.37} & \textbf{3.00}
				& \textbf{45.34} & \textbf{2.75}
				& \textbf{2.66}
				& \textbf{26}
				& \textbf{39} \\
				
				\midrule
				\multicolumn{12}{c}{\textit{Direct zero-shot models}} \\
				\midrule
				TimeRCD
				& \underline{37.00} & \underline{3.88}
				& \underline{83.94} & \underline{3.69}
				& \underline{40.88} & \underline{3.62}
				& \underline{37.03} & \underline{4.19}
				& \underline{3.84}
				& \underline{14}
				& \underline{26} \\
				
				DADA
				& 30.16 & 5.62
				& 81.41 & 4.88
				& 37.29 & 5.84
				& 32.60 & 6.31
				& 5.66
				& 5
				& 10 \\
				
				Chronos
				& 26.66 & 7.09
				& 80.67 & 5.31
				& 32.54 & 6.91
				& 28.52 & 7.12
				& 6.61
				& 2
				& 8 \\
				
				MOMENT
				& 28.57 & 6.81
				& 74.39 & 6.94
				& 25.66 & 7.09
				& 22.31 & 7.50
				& 7.09
				& 0
				& 3 \\
				
				TimesFM
				& 28.59 & 6.44
				& 70.45 & 8.12
				& 32.08 & 7.44
				& 28.71 & 7.44
				& 7.36
				& 1
				& 8 \\
				
				Time-MoE
				& 18.31 & 9.44
				& 69.47 & 9.09
				& 22.29 & 7.81
				& 18.05 & 8.94
				& 8.82
				& 0
				& 0 \\
				
				\midrule
				\multicolumn{12}{c}{\textit{Target-fitted unsupervised models}} \\
				\midrule
				OmniAnomaly
				& 30.73 & 4.75
				& 75.40 & 6.12
				& 34.44 & 4.38
				& 31.60 & 4.44
				& 4.92
				& 10
				& 17 \\
				
				USAD
				& 29.44 & 5.50
				& 68.67 & 7.94
				& 30.83 & 5.28
				& 30.00 & 4.75
				& 5.87
				& 5
				& 13 \\
				
				TranAD
				& 25.84 & 6.25
				& 75.72 & 6.31
				& 25.77 & 7.31
				& 24.60 & 6.69
				& 6.64
				& 1
				& 4 \\
				
				TFMAE
				& 16.59 & 9.28
				& 71.99 & 7.66
				& 19.55 & 9.12
				& 15.81 & 8.56
				& 8.66
				& 0
				& 0 \\
				
				DCdetector
				& 14.99 & 10.50
				& 67.71 & 9.50
				& 16.07 & 10.19
				& 13.27 & 9.31
				& 9.88
				& 0
				& 0 \\
				
				\bottomrule
		\end{tabular}}
        \vspace{-10pt}
	\end{table}
	
	\subsection{Experimental Results}

	\textbf{Overall Performance.}
	Table~\ref{tab:main_results} shows that TS-Router achieves the highest mean detection scores and the best average ranks under all four metrics across the 16 benchmarks. It ranks first in 26 of the 64 dataset--metric combinations and among the top two in 39, indicating benefits across multiple targets and metrics. These results hold against both direct zero-shot models and target-fitted unsupervised detectors, although their deployment protocols differ. In particular, TS-Router reuses the pretrained encoder underlying TimeRCD but uses its representation to estimate specialist competence, followed by target-specific fitting of selected detectors. This comparison demonstrates the effectiveness of the complete routing-and-adaptation pipeline; the controlled analyses below examine how specialist selection and temporal representation contribute to its performance. Appendix~B.2.7 shows that TS-Router leads classical selectors under
	synthetic-to-real and leave-one-dataset-out transfer, while remaining
	competitive when they receive labeled in-distribution support.
	
	\begin{table}[H]
		\caption{
			Routing effectiveness and representation ablations on eleven univariate benchmarks.
			Best Fixed denotes the hindsight-best single specialist shared across targets
			for each evaluation metric, whereas Oracle denotes hindsight per-target
			individual-specialist selection; these are single-specialist detection
			references, not Top-3 fusion oracles.
			Best Top-3 always fuses the hindsight-best set of three specialists
			shared across targets.
			Routing metrics use $k{=}1$ for Best Fixed and Oracle and $k{=}3$ otherwise;
			Router Top-1 shares the TS-Router ranking, so its routing metrics are not repeated.
			Representation variants differ only in the router representation, with the same
			router architecture, specialist pool, supervision, and Top-3 normalized mean fusion.
			The dashed line separates conventional feature representations from frozen
			TSFM encoders (Chronos, MOMENT, TimeRCD);
			TimeRCD corresponds to the default TS-Router configuration.
		}
		\label{tab:routing_effectiveness}
		\centering
		\setlength{\tabcolsep}{4.0pt}
		\renewcommand{\arraystretch}{1.08}
		\begin{tabular}{@{}lcccccc@{}}
			\toprule
			\multirow{2}{*}{Variant}
			& \multicolumn{4}{c}{Detection performance}
			& \multicolumn{2}{c}{Routing quality} \\
			\cmidrule(lr){2-5}
			\cmidrule(lr){6-7}
			& VUS-PR$\uparrow$
			& Aff.-F1$\uparrow$
			& F1$_{\mathrm{T}}$$\uparrow$
			& Std.-F1$\uparrow$
			& NDCG@$k$$\uparrow$
			& Hit@$k$$\uparrow$ \\
			\midrule
			
			\multicolumn{7}{@{}l}{\textit{Selection and fusion strategy}} \\[1pt]
			Best Fixed
			& 39.68 & 84.34 & 41.01 & 36.43
			& 0.607 & 0.207 \\
			
			Full Ensemble
			& 49.54 & 85.55 & 46.22 & 43.48
			& -- & -- \\
			
			Best Fixed Top-3
			& 47.44 & 86.56 & 47.56 & 43.42
			& 0.625 & 0.532 \\
			
			Router Top-1
			& 49.00 & 86.80 & 43.47 & 41.41
			& -- & -- \\
			
			\textbf{TS-Router Top-3}
			& \textbf{56.36}
			& \textbf{88.11}
			& \textbf{49.62}
			& \textbf{47.73}
			& \textbf{0.782}
			& \textbf{0.632} \\
			
			Oracle
			& 68.05 & 94.29 & 66.59 & 64.59
			& 1.000 & 1.000 \\
			
			\midrule
			\multicolumn{7}{@{}l}{\textit{Router representation (Top-3 routing)}} \\[1pt]
			Catch22
			& 49.96 & 84.65 & 42.28 & 40.13
			& 0.674 & 0.426 \\
			
			TSFresh
			& 49.01 & 88.16 & 47.13 & 42.86
			& 0.656 & 0.570 \\
			
			BasicStats
			& 45.48 & 84.32 & 44.60 & 41.31
			& 0.584 & 0.436 \\
			
			MiniRocket
			& 51.31 & 85.91 & 44.63 & 43.53
			& 0.660 & 0.496 \\
			
			\hdashline
			\noalign{\vskip 2pt}
			Chronos
			& 53.27 & 87.25 & 45.93 & 42.79
			& 0.698 & 0.494 \\
			
			MOMENT
			& 53.89
			& \textbf{88.50}
			& \textbf{49.74}
			& 46.37
			& 0.721
			& 0.629 \\
			
			TimeRCD
			& \textbf{56.36}
			& 88.11
			& 49.62
			& \textbf{47.73}
			& \textbf{0.782}
			& \textbf{0.632} \\
			
			\bottomrule
		\end{tabular}
		\vspace{-10pt}
	\end{table}
	
	\textbf{Effectiveness of Competence Routing.}
	The upper block of Table~\ref{tab:routing_effectiveness} examines specialist selection and its downstream use in detection. TS-Router outperforms the hindsight-best fixed individual specialist and full-pool ensembling under all four detection metrics. Best Top-3 always fuses the hindsight-best set of three specialists shared across targets, yet remains below TS-Router Top-3 on both detection and routing metrics, supporting target-adaptive selection over that globally fixed set. The NDCG and Hit scores assess agreement with individual specialist competence before fusion. Top-3 improves over Router Top-1 under all four detection metrics, and TS-Router's advantage over full-pool ensembling supports selective rather than indiscriminate fusion. These comparisons evaluate the detection effect of combining the selected score sequences, which is separate from the fixed-budget set-competence analysis.
	
	\textbf{Role of the Foundation Representation.}
	The lower block of Table~\ref{tab:routing_effectiveness} examines the representation supporting this competence estimation, holding the router architecture, specialist pool, supervision, and fusion procedure fixed. All three frozen TSFM encoders achieve higher VUS-PR and NDCG than the conventional feature representations. TimeRCD provides the highest NDCG and Hit scores, together with the best VUS-PR and Standard-F1, while MOMENT achieves the highest Affiliation-F1 and F1$_{\mathrm{T}}$. The results with Chronos and MOMENT show that effective foundation-guided routing is not confined to the default TimeRCD encoder. Differences across metrics also indicate that stronger ranking quality need not yield uniformly better detection scores. Taken together, the two blocks support the representation-to-competence-to-detection pathway: pretrained representations inform specialist suitability, competence-based routing identifies useful subsets, and selective fusion improves detection. Consistent with Section~\ref{sec:problem}, these findings support the generalist-representation, specialist-detection perspective: transferable temporal knowledge guides which criteria to trust, while selected specialists perform anomaly detection.

	\section{Conclusion}
	
	We introduced TS-Router, a generalist-representation, specialist-detection framework that reconsiders the role of foundation models in time-series anomaly detection. TS-Router uses transferable temporal representations to estimate specialist competence and select suitable detectors, with labeled simulation providing soft routing supervision without requiring large-scale real anomaly annotations. Our analysis bounds Top-\(k\) set-competence regret under explicit transfer assumptions, while experiments evaluate the complete representation-to-competence-to-detection pathway: foundation representations inform specialist suitability, and selective score fusion improves detection. Results with multiple frozen TSFM encoders further show that this role is not confined to a single pretrained model. As with other simulation-based approaches, competence transfer is shaped by
	how well the simulated task distribution covers variations relevant to
	unseen targets. For multivariate series, we focus on variable-wise routing
	to isolate the representation-to-competence-to-detection pathway;
	incorporating cross-variable interactions is a natural extension. 
	\clearpage
	
	
\subsection*{AI Use Statement}

Generative AI tools were used during the research and manuscript preparation
for brainstorming and discussing alternative formulations of the research
problem, providing feedback on experimental design and interpretation of
empirical results, assisting with drafting, restructuring, and polishing
parts of the manuscript, and serving as a checking and discussion aid for
the theoretical component. For mathematical arguments and proofs,
generative AI was used primarily to inspect derivations, identify possible
inconsistencies or unclear assumptions, and help assess whether proof steps
and stated conclusions were logically aligned, rather than to replace the
authors' own mathematical reasoning. The final TS-Router formulation,
assumptions, proofs, experimental procedures, reported results, and
conclusions were independently examined and determined by the authors.
Generative AI was not used to discover or retrieve related literature or to
generate the simulated datasets used in the experiments. All AI-assisted
suggestions were critically reviewed before inclusion, and the authors take
full responsibility for the final content of this work, including its text,
mathematical claims, experimental results, and conclusions.

\subsection*{Ethics Statement}

This work studies time-series anomaly detection using public real-world
benchmark datasets and simulated Syn-RCD tasks. It does not involve human
subjects, personally identifiable information, or interventions in
real-world systems. The method is intended for research on anomaly
detection; deployment in safety-critical domains requires domain-specific
validation and appropriate human oversight.

\subsection*{Reproducibility Statement}

We provide detailed descriptions of TS-Router's competence-construction,
router-training, and target-deployment procedures in Section~4,
Algorithm~\ref{alg:tsrouter}, and Appendix~B. Dataset descriptions,
preprocessing, specialist definitions and hyperparameters, baseline
configurations, evaluation metrics, and implementation settings are
reported in Appendix~B. Additional per-dataset results, representation
ablations, specialist complementarity analyses, Top-$k$ and specialist-pool
sensitivity, routing-head ablations, and algorithm-selection transfer
experiments are also provided in Appendix~B. Complete assumptions and proofs
of the theoretical results are given in Appendix~A. The supplementary
materials provide the Syn-RCD simulation specifications, competence-target
construction details, and implementation resources needed to reproduce the
experiments. The code is available at \url{https://anonymous.4open.science/r/TS-Router-D8FF}.
	
	\bibliography{iclr2027_conference}
	\bibliographystyle{iclr2027_conference}
	
	\appendix
	\clearpage
	
	\section{Additional Methodological Details}
	\label{app:method_details}
	
	\subsection{Theoretical Supplement}
	\label{app:proof}
	
	This section proves the budget-\(k\) regret and transfer results, establishes the associated utility hierarchy, and relates specialist-set recovery to score fusion. Throughout, \(K\ge2\), \(1\le k\le K\), \(\tau>0\), and \(\mathbf c(\mathcal T)\in[0,1]^K\). The encoder, pooling operation, specialist pool, and label-free fitting and scoring protocol are fixed. Task and representation spaces are standard Borel spaces, so regular conditional distributions exist. All conditional statements hold almost surely under the relevant representation distribution. The router \(\mathbf p_\theta(z)=\hat{\mathbf q}_\theta(z)\) is measurable and strictly positive. Every Top-\(k\) operation uses the same fixed measurable tie-breaking rule and returns a set of indices in \(\mathcal I_k\). We use the notation \(V_k\), \(\mathsf T_k\), \(m_k\), \(\alpha_k\), and \(C_{\tau,K,k}\) from Sections~\ref{sec:problem} and~\ref{sec:theory}.
	
	\subsubsection{Proof of Lemma~1}
	\label{app:proof_lemma1}
	
	\textit{Equal-budget set differences.}
	For any \(S,S'\in\mathcal I_k\), the sets \(A=S\setminus S'\) and \(B=S'\setminus S\) have the same cardinality \(r\le\min\{k,K-k\}=m_k\), since \(|S|=|S'|=k\). Canceling the common members \(S\cap S'\) and pairing the remaining indices by any bijection \(\pi:A\to B\) gives, for every \(\mathbf u\in\mathbb R^K\),
	\[
	V_k(S,\mathbf u)-V_k(S',\mathbf u)
	=\frac1k\sum_{i\in A}\bigl(u_i-u_{\pi(i)}\bigr).
	\]
	Each summand lies in \([-\operatorname{osc}(\mathbf u),\operatorname{osc}(\mathbf u)]\), so
	\begin{equation}
		|V_k(S,\mathbf u)-V_k(S',\mathbf u)|
		\le\frac{r}{k}\operatorname{osc}(\mathbf u)
		\le\alpha_k\operatorname{osc}(\mathbf u).
		\label{eq:app_set_perturbation}
	\end{equation}
	In particular, every selection regret for a vector of range at most one is bounded by \(\alpha_k\). If \(k=K\), then \(\mathcal I_K\) is a singleton, \(m_K=0\), and both the regret and the second term in \(C_{\tau,K,K}\) vanish, so \(C_{\tau,K,K}=0\). Henceforth take \(k<K\).
	
	\textit{Probability mass at the selection boundary.}
	Let \(v_{(1)}\ge\cdots\ge v_{(K)}\) denote the ordered coordinates of \(\mathbf v\), and let \(q_{(k)}\) be the \(k\)-th largest coordinate of \(\mathbf q=\operatorname{Softmax}(\mathbf v/\tau)\). Softmax is invariant under subtracting \(v_{(k)}\), so
	\[
	q_{(k)}
	=\frac{e^{v_{(k)}/\tau}}{\sum_{\ell=1}^K e^{v_{(\ell)}/\tau}}
	=\left(\sum_{\ell=1}^K e^{(v_{(\ell)}-v_{(k)})/\tau}\right)^{-1}.
	\]
	Since \(\operatorname{osc}(\mathbf v)\le1\), one has \(v_{(\ell)}-v_{(k)}\le1\) for \(\ell<k\) and \(v_{(\ell)}-v_{(k)}\le0\) for \(\ell>k\). Thus the first \(k-1\) summands are at most \(e^{1/\tau}\), the remaining \(K-k+1\) summands (including \(\ell=k\)) are at most one, and
	\begin{equation}
		q_{(k)}
		\ge\frac1{(k-1)e^{1/\tau}+K-k+1}
		=\frac1{D_{\tau,K,k}}.
		\label{eq:app_boundary_mass}
	\end{equation}
	
	Let \(S^*=\operatorname{TopK}(\mathbf v,k)\) and \(\widehat S=\operatorname{TopK}(\mathbf p,k)\). If the two sets agree, the regret is zero. Otherwise, write \(A=S^*\setminus\widehat S\), \(B=\widehat S\setminus S^*\), and \(|A|=|B|=r\ge1\). Choose any bijection \(\pi:A\to B\). For \(i\in A\), put \(d_i=v_i-v_{\pi(i)}\). Because \(S^*\) maximises \(V_k(\cdot,\mathbf v)\), every \(i\in S^*\) and \(j\notin S^*\) satisfy \(v_i\ge v_j\), hence \(d_i\ge0\). The range bound \(\operatorname{osc}(\mathbf v)\le1\) gives \(d_i\le1\). Softmax preserves the order of \(\mathbf v\) and the same tie-breaking rule, so \(S^*=\operatorname{TopK}(\mathbf q,k)\) and therefore \(q_i\ge q_{(k)}\ge D_{\tau,K,k}^{-1}\). The softmax ratios yield \(q_{\pi(i)}/q_i=e^{-d_i/\tau}\). Because \(\widehat S\) maximises \(V_k(\cdot,\mathbf p)\), one has \(p_{\pi(i)}\ge p_i\). Collecting these relations,
	\begin{equation}
		0\le d_i\le1,\qquad
		q_i\ge D_{\tau,K,k}^{-1},\qquad
		q_{\pi(i)}=q_i e^{-d_i/\tau},\qquad
		p_{\pi(i)}\ge p_i.
		\label{eq:app_exchange_pairs}
	\end{equation}
	Canceling common members of \(S^*\) and \(\widehat S\) gives
	\[
	\mathsf T_k(\mathbf v)-V_k(\widehat S,\mathbf v)
	=\frac1k\sum_{i\in A}d_i.
	\]
	
	\textit{Probability differences.}
	The map \(x\mapsto1-e^{-x/\tau}\) is concave, so on \([0,1]\) it lies above the chord joining \(0\) and \(1\): \(1-e^{-d_i/\tau}\ge d_i(1-e^{-1/\tau})\). Combined with \(q_i-q_{\pi(i)}=q_i(1-e^{-d_i/\tau})\) and \(q_i\ge D_{\tau,K,k}^{-1}\),
	\[
	q_i-q_{\pi(i)}
	\ge\frac{1-e^{-1/\tau}}{D_{\tau,K,k}}d_i.
	\]
	For the comparison with \(\mathbf p\), write
	\[
	q_i-q_{\pi(i)}
	=(q_i-p_i)+(p_i-p_{\pi(i)})+(p_{\pi(i)}-q_{\pi(i)}).
	\]
	The middle term is nonpositive by Eq.(~\ref{eq:app_exchange_pairs}), so
	\[
	q_i-q_{\pi(i)}
	\le|q_i-p_i|+|q_{\pi(i)}-p_{\pi(i)}|.
	\]
	The pairs \((i,\pi(i))\) are disjoint. Summing therefore yields
	\begin{align}
		\frac{1-e^{-1/\tau}}{D_{\tau,K,k}}\sum_{i\in A}d_i
		&\le\sum_{i\in A}(q_i-q_{\pi(i)})\nonumber\\
		&\le\sum_{j\in A\cup B}|q_j-p_j|
		\le\|\mathbf q-\mathbf p\|_1.
		\label{eq:app_probability_gap}
	\end{align}
	Pinsker's inequality states that \(\|\mathbf q-\mathbf p\|_1\le\sqrt{2D_{\mathrm{KL}}(\mathbf q\Vert\mathbf p)}\). To verify the constant, when \(\mathbf q\ne\mathbf p\), let \(I=\{j:q_j\ge p_j\}\), \(a=\sum_{j\in I}q_j\), and \(b=\sum_{j\in I}p_j\). Then \(a\ge b\) and \(\|\mathbf q-\mathbf p\|_1=2(a-b)\). Since \(\mathbf p\) is strictly positive, \(b\in(0,1)\). The log-sum inequality gives
	\[
	D_{\mathrm{KL}}(\mathbf q\Vert\mathbf p)\ge\operatorname{kl}(a\Vert b),
	\qquad
	\operatorname{kl}(a\Vert b)=a\log\frac ab+(1-a)\log\frac{1-a}{1-b}.
	\]
	For fixed \(b\in(0,1)\), this binary divergence and its first derivative vanish at \(a=b\), while its second derivative is \(1/[a(1-a)]\ge4\). Taylor expansion with remainder therefore yields \(\operatorname{kl}(a\Vert b)\ge2(a-b)^2=\tfrac12\|\mathbf q-\mathbf p\|_1^2\). The case \(\mathbf q=\mathbf p\) is immediate. Applying this inequality to Eq.(~\ref{eq:app_probability_gap}) yields
	\begin{equation}
		\frac1k\sum_{i\in A}d_i
		\le\frac{D_{\tau,K,k}}{k(1-e^{-1/\tau})}
		\sqrt{2D_{\mathrm{KL}}(\mathbf q\Vert\mathbf p)}.
		\label{eq:app_probability_regret}
	\end{equation}
	
	\textit{Square-root probability differences.}
	The same chord bound at temperature \(2\tau\) and \(q_i\ge D_{\tau,K,k}^{-1}\) give
	\[
	\sqrt{q_i}-\sqrt{q_{\pi(i)}}
	=\sqrt{q_i}(1-e^{-d_i/(2\tau)})
	\ge\frac{1-e^{-1/(2\tau)}}{\sqrt{D_{\tau,K,k}}}d_i.
	\]
	Because \(p_{\pi(i)}\ge p_i\),
	\[
	\sqrt{q_i}-\sqrt{q_{\pi(i)}}
	\le\bigl|\sqrt{q_i}-\sqrt{p_i}\bigr|
	+\bigl|\sqrt{q_{\pi(i)}}-\sqrt{p_{\pi(i)}}\bigr|.
	\]
	Summing over the \(r\) disjoint pairs and applying Cauchy--Schwarz on the \(2r\) coordinates in \(A\cup B\) yields
	\begin{equation}
		\frac{1-e^{-1/(2\tau)}}{\sqrt{D_{\tau,K,k}}}\sum_{i\in A}d_i
		\le\sqrt{2r}\,H(\mathbf q,\mathbf p),
		\qquad
		H^2(\mathbf q,\mathbf p)=\sum_{j=1}^K(\sqrt{q_j}-\sqrt{p_j})^2.
		\label{eq:app_root_gap}
	\end{equation}
	Let \(a(\mathbf q,\mathbf p)=\sum_j\sqrt{q_jp_j}\). Then \(H^2(\mathbf q,\mathbf p)=2-2a(\mathbf q,\mathbf p)\). Jensen's inequality for the concave logarithm and \(-\log x\ge1-x\) imply
	\begin{equation}
		D_{\mathrm{KL}}(\mathbf q\Vert\mathbf p)
		=-2\sum_jq_j\log\sqrt{\frac{p_j}{q_j}}
		\ge-2\log a(\mathbf q,\mathbf p)
		\ge2\bigl(1-a(\mathbf q,\mathbf p)\bigr)
		=H^2(\mathbf q,\mathbf p).
		\label{eq:app_hellinger_kl}
	\end{equation}
	Combining Eqs.(~\ref{eq:app_root_gap}) and~(~\ref{eq:app_hellinger_kl}) and using \(H\le\sqrt{D_{\mathrm{KL}}(\mathbf q\Vert\mathbf p)}\) together with \(r\le m_k\) gives
	\begin{equation}
		\frac1k\sum_{i\in A}d_i
		\le\frac{\sqrt{m_kD_{\tau,K,k}}}{k(1-e^{-1/(2\tau)})}
		\sqrt{2D_{\mathrm{KL}}(\mathbf q\Vert\mathbf p)}.
		\label{eq:app_root_regret}
	\end{equation}
	Taking the smaller of Eqs.(~\ref{eq:app_probability_regret}) and~(~\ref{eq:app_root_regret}), together with the bound \(\alpha_k\) from Eq.(~\ref{eq:app_set_perturbation}), proves Lemma~1.
	\hfill\(\square\)
	
	\subsubsection{Proof of Theorem~1}
	\label{app:proof_theorem1}
	
	\textit{Population target of the KL objective.}
	For fixed \(z\), abbreviate \(\mathbf q=\mathbf q(\mathcal T)\), \(\mathbf m=\mathbf m_s(z)\), and \(\mathbf p=\mathbf p_\theta(z)\). Since \(\mathbf p\) and \(\mathbf m\) are functions of \(z\) and \(\mathbb E_{P_s}[q_j\mid Z=z]=m_j\), splitting \(\log(q_j/p_j)=\log(q_j/m_j)+\log(m_j/p_j)\) gives
	\[
	\mathbb E_{P_s}\!\left[\sum_{j=1}^K q_j\log\frac{q_j}{p_j}\,\middle|\,Z=z\right]
	=\mathbb E_{P_s}\!\left[\sum_{j=1}^K q_j\log\frac{q_j}{m_j}\,\middle|\,Z=z\right]
	+\sum_{j=1}^K m_j\log\frac{m_j}{p_j},
	\]
	which is
	\begin{equation}
		\mathbb E_{P_s}[D_{\mathrm{KL}}(\mathbf q\Vert\mathbf p)\mid Z=z]
		=\mathbb E_{P_s}[D_{\mathrm{KL}}(\mathbf q\Vert\mathbf m)\mid Z=z]
		+D_{\mathrm{KL}}(\mathbf m\Vert\mathbf p).
		\label{eq:app_conditional_risk_identity}
	\end{equation}
	Because \(\mathbf c(\mathcal T)\in[0,1]^K\),
	\[
	q_j(\mathcal T)
	=\frac{e^{c_j(\mathcal T)/\tau}}{\sum_{\ell=1}^K e^{c_\ell(\mathcal T)/\tau}}
	\ge K^{-1}e^{-1/\tau}>0,
	\]
	and the same lower bound passes to \(\mathbf m_s\) by conditional expectation. The target-dependent logarithms are therefore bounded, and finite \(\mathcal L_s(\theta)\) justifies integration. The second term in Eq.(~\ref{eq:app_conditional_risk_identity}) is minimized at the measurable predictor \(\mathbf p(z)=\mathbf m_s(z)\). Therefore
	\begin{align}
		\mathcal L_s^*
		&=\mathbb E_{P_s}D_{\mathrm{KL}}(\mathbf q(\mathcal T)\Vert\mathbf m_s(Z)),
		\label{eq:app_irreducible_risk}\\
		\mathcal E_s(\theta)
		&=\mathbb E_{P_s^Z}D_{\mathrm{KL}}(\mathbf m_s(Z)\Vert\mathbf p_\theta(Z)).
		\label{eq:app_excess_identity}
	\end{align}
	This infimum is over all measurable simplex-valued predictors, not only the chosen router architecture.
	
	\textit{Range of the conditional score vector.}
	Write \(\mathbf h_s(z)=\tau\log\mathbf m_s(z)\), with the logarithm taken coordinatewise. Since \(|c_i-c_j|\le1\), each task satisfies
	\[
	e^{-1/\tau}q_j(\mathcal T)\le q_i(\mathcal T)\le e^{1/\tau}q_j(\mathcal T).
	\]
	These are linear inequalities in \(\mathbf q(\mathcal T)\), so they pass to the conditional means: \(e^{-1/\tau}m_{s,j}(z)\le m_{s,i}(z)\le e^{1/\tau}m_{s,j}(z)\). Taking logarithms yields \(|\tau\log(m_{s,i}(z)/m_{s,j}(z))|\le1\). Consequently,
	\begin{equation}
		\operatorname{osc}(\mathbf h_s(z))\le1,
		\qquad
		\operatorname{Softmax}(\mathbf h_s(z)/\tau)=\mathbf m_s(z).
		\label{eq:app_effective_competence}
	\end{equation}
	Lemma~1 thus applies to \(\mathbf h_s(z)\), even though the conditional mean soft target need not be the softmax of the conditional mean competence. Also, \(0\le b_{\tau,s}(z)\le\operatorname{osc}(\boldsymbol\mu_s(z))+\operatorname{osc}(\mathbf h_s(z))\le2\), so the distortion is integrable.
	
	\textit{Conditional Top-\(k\) regret.}
	For \(d\in\{s,r\}\), define
	\begin{equation}
		R_{d,k}(z)=\mathsf T_k(\boldsymbol\mu_d(z))
		-V_k(\widehat S_\theta^{(k)}(z),\boldsymbol\mu_d(z)).
		\label{eq:app_cond_regret}
	\end{equation}
	Fix \(z\), set \(\widehat S=\widehat S_\theta^{(k)}(z)\), and let \(S_s^*=\operatorname{TopK}(\boldsymbol\mu_s(z),k)\). Write \(\mathbf e_s=\boldsymbol\mu_s-\mathbf h_s\), so that \(b_{\tau,s}(z)=\operatorname{osc}(\mathbf e_s(z))\) by Eq.(~\ref{eq:conditional_softmax_distortion}). Then
	\[
	R_{s,k}(z)
	=V_k(S_s^*,\boldsymbol\mu_s(z))-V_k(\widehat S,\boldsymbol\mu_s(z))
	=V_k(S_s^*,\mathbf h_s(z))-V_k(\widehat S,\mathbf h_s(z))
	+V_k(S_s^*,\mathbf e_s(z))-V_k(\widehat S,\mathbf e_s(z)).
	\]
	The first difference is at most \(\mathsf T_k(\mathbf h_s(z))-V_k(\widehat S,\mathbf h_s(z))\). The second is at most \(\alpha_k b_{\tau,s}(z)\) by Eq.(~\ref{eq:app_set_perturbation}). Lemma~1 applied to \(\mathbf h_s(z)\), using Eq.(~\ref{eq:app_effective_competence}), therefore gives
	\begin{equation}
		R_{s,k}(z)
		\le C_{\tau,K,k}\sqrt{2D_{\mathrm{KL}}(\mathbf m_s(z)\Vert\mathbf p_\theta(z))}
		+\alpha_k b_{\tau,s}(z).
		\label{eq:app_source_excess_regret}
	\end{equation}
	
	There is also a bound directly in terms of the total routing loss. Let
	\[
	\eta_s(z)=\mathbb E_{P_s}[D_{\mathrm{KL}}(\mathbf q(\mathcal T)\Vert\mathbf p_\theta(z))\mid Z=z].
	\]
	The function \(\mathsf T_k\) is convex, being a maximum of finitely many linear functions, and \(\widehat S\) is fixed conditional on \(Z=z\). Jensen's inequality and linearity of \(V_k(\widehat S,\cdot)\) therefore yield
	\[
	R_{s,k}(z)
	=\mathsf T_k(\boldsymbol\mu_s(z))-V_k(\widehat S,\boldsymbol\mu_s(z))
	\le\mathbb E_{P_s}\!\left[
	\mathsf T_k(\mathbf c(\mathcal T))-V_k(\widehat S,\mathbf c(\mathcal T))
	\,\middle|\,Z=z\right].
	\]
	Lemma~1 applies task-wise because \(\operatorname{osc}(\mathbf c(\mathcal T))\le1\). Concavity of the square root then gives
	\begin{align}
		R_{s,k}(z)
		&\le C_{\tau,K,k}\,\mathbb E_{P_s}\!\left[
		\sqrt{2D_{\mathrm{KL}}(\mathbf q(\mathcal T)\Vert\mathbf p_\theta(z))}
		\,\middle|\,Z=z\right]\nonumber\\
		&\le C_{\tau,K,k}\sqrt{2\eta_s(z)}.
		\label{eq:app_source_total_regret}
	\end{align}
	The tower property gives \(\mathbb E_{P_s^Z}\eta_s(Z)=\mathcal L_s(\theta)\).
	
	\textit{Conditional competence shift.}
	Write \(\mathbf d(z)=\boldsymbol\mu_r(z)-\boldsymbol\mu_s(z)\) and \(S_r^*=\operatorname{TopK}(\boldsymbol\mu_r(z),k)\). Then
	\begin{align*}
		R_{r,k}(z)
		&=V_k(S_r^*,\boldsymbol\mu_s(z)+\mathbf d(z))
		-V_k(\widehat S,\boldsymbol\mu_s(z)+\mathbf d(z))\\
		&=\bigl(V_k(S_r^*,\boldsymbol\mu_s(z))-V_k(\widehat S,\boldsymbol\mu_s(z))\bigr)
		+\bigl(V_k(S_r^*,\mathbf d(z))-V_k(\widehat S,\mathbf d(z))\bigr).
	\end{align*}
	The first difference is at most \(R_{s,k}(z)\). The second is at most \(\alpha_k\operatorname{osc}(\mathbf d(z))\) by Eq.(~\ref{eq:app_set_perturbation}). The assumption \(\|\boldsymbol\mu_r-\boldsymbol\mu_s\|_\infty\le\delta\) gives \(\operatorname{osc}(\mathbf d(z))\le2\delta\), hence
	\begin{equation}
		R_{r,k}(z)
		\le R_{s,k}(z)+\alpha_k\operatorname{osc}(\mathbf d(z))
		\le R_{s,k}(z)+2\alpha_k\delta.
		\label{eq:app_domain_shift_regret}
	\end{equation}
	
	\textit{Transfer across representation distributions.}
	Let \(w=dP_r^Z/dP_s^Z\). Then \(0\le w\le\rho\), so \(w^2\le\rho w\). For any nonnegative measurable \(f\) with finite source expectation, the Cauchy--Schwarz inequality \(\mathbb E[w\sqrt{f}]\le\sqrt{\mathbb E[w^2]\,\mathbb E[f]}\) and \(\mathbb E_{P_s^Z}w=1\) imply
	\begin{equation}
		\mathbb E_{P_r^Z}\sqrt{f(Z)}
		=\mathbb E_{P_s^Z}[w(Z)\sqrt{f(Z)}]
		\le\sqrt{\mathbb E_{P_s^Z}[w(Z)^2]\,\mathbb E_{P_s^Z}[f(Z)]}
		\le\sqrt{\rho\,\mathbb E_{P_s^Z}[f(Z)]}.
		\label{eq:app_density_ratio}
	\end{equation}
	Absolute continuity transfers source-almost-sure conditional bounds to \(P_r^Z\). Taking target expectations in Eq.(~\ref{eq:app_domain_shift_regret}) and using Eq.(~\ref{eq:app_source_excess_regret}) gives
	\begin{align}
		\mathbb E_{P_r^Z}R_{r,k}(Z)
		&\le C_{\tau,K,k}\mathbb E_{P_r^Z}
		\sqrt{2D_{\mathrm{KL}}(\mathbf m_s(Z)\Vert\mathbf p_\theta(Z))}
		+\alpha_k\beta_{\tau,r}+2\alpha_k\delta\nonumber\\
		&\le C_{\tau,K,k}\sqrt{2\rho\,\mathcal E_s(\theta)}
		+\alpha_k(\beta_{\tau,r}+2\delta).
		\label{eq:app_target_excess_regret}
	\end{align}
	Using Eq.(~\ref{eq:app_source_total_regret}) instead, together with \(\mathbb E_{P_s^Z}\eta_s(Z)=\mathcal L_s(\theta)\), gives
	\begin{equation}
		\mathbb E_{P_r^Z}R_{r,k}(Z)
		\le C_{\tau,K,k}\sqrt{2\rho\,\mathcal L_s(\theta)}+2\alpha_k\delta.
		\label{eq:app_target_total_regret}
	\end{equation}
	By definition, \(\mathbb E_{P_r^Z}R_{r,k}(Z)=U_{\mathrm{cond},r}^{(k)}(Z)-U_{\mathrm{router},r}^{(k)}\). Each coordinate of \(\boldsymbol\mu_r(z)\) lies in \([0,1]\), so \(\operatorname{osc}(\boldsymbol\mu_r(z))\le1\) and Eq.(~\ref{eq:app_set_perturbation}) yields \(0\le R_{r,k}(z)\le\alpha_k\). Combining this trivial bound with Eqs.(~\ref{eq:app_target_excess_regret}) and~(~\ref{eq:app_target_total_regret}) proves both inequalities in Theorem~1, including the case \(k=K\), where \(\alpha_K=C_{\tau,K,K}=0\).
	\hfill\(\square\)
	
	\subsubsection{Conditional Softmax Distortion}
	\label{app:conditional_distortion}
	
	The distortion in Eq.(~\ref{eq:conditional_softmax_distortion}) measures differences between conditional competence gaps and log-ratios of conditional mean soft targets. For fixed \(P_s\), \(E\), and \(\tau\), it is independent of the router parameters.
	
	\textit{Coordinate-wise Jensen gaps.}
	For \(j\in[K]\), define
	\begin{equation}
		J_j(z)=\log\mathbb E_{P_s}[q_j(\mathcal T)\mid Z=z]
		-\mathbb E_{P_s}[\log q_j(\mathcal T)\mid Z=z]\ge0.
		\label{eq:app_jensen_gap}
	\end{equation}
	Let \(F_\tau(\mathbf c)=\tau\log\sum_{\ell=1}^K e^{c_\ell/\tau}\). Then \(\tau\log q_j=c_j-F_\tau(\mathbf c)\), so
	\[
	\mathbb E_{P_s}[\tau\log q_j(\mathcal T)\mid Z=z]
	=\mu_{s,j}(z)-\mathbb E_{P_s}[F_\tau(\mathbf c(\mathcal T))\mid Z=z].
	\]
	The definition of \(J_j\) rearranges to \(\tau\log m_{s,j}(z)=\mathbb E_{P_s}[\tau\log q_j(\mathcal T)\mid Z=z]+\tau J_j(z)\), hence
	\[
	h_{s,j}(z)=\mu_{s,j}(z)-\mathbb E_{P_s}[F_\tau(\mathbf c(\mathcal T))\mid Z=z]+\tau J_j(z).
	\]
	The conditional log-normalizer does not depend on \(j\), so
	\[
	\boldsymbol\mu_s(z)-\mathbf h_s(z)
	=\mathbb E_{P_s}[F_\tau(\mathbf c(\mathcal T))\mid Z=z]\,\mathbf 1
	-\tau\mathbf J(z)
	\]
	and therefore
	\begin{equation}
		b_{\tau,s}(z)=\operatorname{osc}(\boldsymbol\mu_s(z)-\mathbf h_s(z))
		=\tau\operatorname{osc}(\mathbf J(z)).
		\label{eq:app_distortion_identity}
	\end{equation}
	A common additive gap across specialists does not change \(\operatorname{osc}(\mathbf J)\).
	
	\textit{A sufficient condition for zero distortion.}
	Suppose that, under \(P_s\),
	\begin{equation}
		\mathbf c(\mathcal T)=\mathbf v(Z)+a(\mathcal T)\mathbf1,
		\label{eq:app_relative_sufficiency}
	\end{equation}
	where \(\mathbf v\) is measurable and \(a\) is integrable. This means that \(Z\) determines all pairwise competence differences, while overall competence levels may vary through a common task-dependent offset. Softmax is invariant under such offsets, so \(\mathbf q(\mathcal T)=\operatorname{Softmax}(\mathbf v(Z)/\tau)\) is conditionally deterministic. Hence \(\mathbf J(z)=\mathbf0\), \(b_{\tau,s}(z)=0\), and absolute continuity gives \(\beta_{\tau,r}=0\). Conversely, if all differences \(c_j-c_1\) are measurable functions of \(Z\), Eq.(~\ref{eq:app_relative_sufficiency}) holds with \(v_1=0\), \(v_j(Z)=c_j-c_1\), and \(a=c_1\).
	
	For fixed \(K,k,\tau,\rho\), a sequence of routers with \(\mathcal E_s(\theta_n)\to0\), together with \(\beta_{\tau,r}=0\) and \(\delta=0\), therefore satisfies
	\begin{equation}
		U_{\mathrm{cond},r}^{(k)}(Z)-U_{\mathrm{router},r}^{(k)}(\theta_n)\longrightarrow0.
		\label{eq:app_conditional_consistency}
	\end{equation}
	When distortion is nonzero, it remains explicit in the bound.
	
	\textit{A quantitative bound under conditional concentration.}
	Suppose that, under \(P_s\) conditional on \(Z=z\), each \(Y_j=\tau\log q_j(\mathcal T)\) lies in an interval of length at most \(r(z)\), where \(r\) is measurable and \(\mathbb E_{P_r^Z}[r(Z)^2]<\infty\). Then
	\begin{equation}
		b_{\tau,s}(z)\le\frac{r(z)^2}{8\tau},
		\qquad
		\beta_{\tau,r}\le\frac{\mathbb E_{P_r^Z}[r(Z)^2]}{8\tau}.
		\label{eq:app_distortion_concentration}
	\end{equation}
	To prove this, fix \(z,j\), write \(Y=Y_j\), \(r=r(z)\), and set \(G(t)=\log\mathbb E_{P_s}[e^{tY}\mid Z=z]\). Then \(G(0)=0\) and \(G'(0)=\mathbb E_{P_s}[Y\mid Z=z]\). Under exponential tilting, \(G''(t)\) is the variance of a random variable supported on an interval of length \(r\). If \(m\) denotes the midpoint of that interval, \(\mathbb E[(Y-m)^2]\le(r/2)^2=r^2/4\), and the variance is at most this second moment, so \(G''(t)\le r^2/4\). The integral remainder
	\[
	G(t)-tG'(0)=\int_0^t(t-u)G''(u)\,du\le\frac{r^2t^2}{8},\qquad t\ge0,
	\]
	together with \(J_j(z)=G(1/\tau)-(1/\tau)G'(0)\), yields \(0\le J_j(z)\le r(z)^2/(8\tau^2)\). Therefore \(\tau\operatorname{osc}(\mathbf J(z))\le\tau\max_jJ_j(z)\le r(z)^2/(8\tau)\), proving Eq.(~\ref{eq:app_distortion_concentration}) after target averaging.
	
	\subsubsection{Utility Hierarchy and Representation Information}
	\label{app:theory_discussion}
	
	\textit{Fixed, conditional, and hindsight selection.}
	Under any task distribution, write \(\boldsymbol\mu(Z)=\mathbb E[\mathbf c(\mathcal T)\mid Z]\). Linearity of \(V_k(S,\cdot)\) and the tower property give \(\mathbb E[V_k(S,\mathbf c(\mathcal T))]=\mathbb E_Z[V_k(S,\boldsymbol\mu(Z))]\) for every fixed \(S\in\mathcal I_k\). Maximising over \(S\) and using \(\mathsf T_k(\boldsymbol\mu(Z))=\max_S V_k(S,\boldsymbol\mu(Z))\) yields
	\begin{align}
		U_{\mathrm{fixed}}^{(k)}
		&=\max_{S\in\mathcal I_k}\mathbb E_Z[V_k(S,\boldsymbol\mu(Z))]
		\le\mathbb E_Z[\mathsf T_k(\boldsymbol\mu(Z))]
		=U_{\mathrm{cond}}^{(k)}(Z),
		\label{eq:app_fixed_cond}\\
		U_{\mathrm{cond}}^{(k)}(Z)
		&=\mathbb E_Z[\mathsf T_k(\mathbb E[\mathbf c(\mathcal T)\mid Z])]
		\le\mathbb E_Z\mathbb E[\mathsf T_k(\mathbf c(\mathcal T))\mid Z]
		=U_{\mathrm{oracle}}^{(k)},
		\label{eq:app_cond_oracle}
	\end{align}
	where the second comparison is Jensen's inequality for the convex map \(\mathsf T_k\). The conditional optimum is attained by the measurable set \(\operatorname{TopK}(\boldsymbol\mu(Z),k)\), establishing Eq.(~\ref{eq:utility_hierarchy}).
	
	If \(Z_0=h(Z_1)\) for a measurable \(h\), then \(\mathbb E[\mathbf c\mid Z_0]=\mathbb E[\mathbb E[\mathbf c\mid Z_1]\mid Z_0]\). Conditional Jensen for \(\mathsf T_k\) gives
	\[
	\mathsf T_k(\boldsymbol\mu(Z_0))
	\le\mathbb E\bigl[\mathsf T_k(\boldsymbol\mu(Z_1))\bigm|Z_0\bigr],
	\]
	and taking expectations yields
	\begin{equation}
		U_{\mathrm{cond}}^{(k)}(Z_0)
		\le\mathbb E\!\left[\mathbb E[\mathsf T_k(\mathbb E[\mathbf c\mid Z_1])\mid Z_0]\right]
		=U_{\mathrm{cond}}^{(k)}(Z_1).
		\label{eq:app_rep_monotonicity}
	\end{equation}
	This monotonicity concerns the best attainable conditional competence at the same budget, not the performance of every trained router.
	
	\textit{Representation and learned selection.}
	Because \(\widehat S_\theta^{(k)}(Z)\) is \(Z\)-measurable, the tower property gives
	\[
	U_{\mathrm{router},r}^{(k)}
	=\mathbb E_{P_r}[V_k(\widehat S_\theta^{(k)}(Z),\mathbf c(\mathcal T))]
	=\mathbb E_{P_r^Z}[V_k(\widehat S_\theta^{(k)}(Z),\boldsymbol\mu_r(Z))].
	\]
	Adding and subtracting \(U_{\mathrm{cond},r}^{(k)}(Z)\) therefore yields
	\begin{align*}
		U_{\mathrm{oracle},r}^{(k)}-U_{\mathrm{router},r}^{(k)}
		&=\Delta_{\mathrm{rep}}^{(k)}(Z)+\mathbb E_{P_r^Z}R_{r,k}(Z),\\
		U_{\mathrm{router},r}^{(k)}-U_{\mathrm{fixed},r}^{(k)}
		&=\Gamma_{\mathrm{route}}^{(k)}(Z)-\mathbb E_{P_r^Z}R_{r,k}(Z).
	\end{align*}
	Theorem~1 supplies \(\mathbb E_{P_r^Z}R_{r,k}(Z)\le\epsilon_k(\theta)\), which is Eq.(~\ref{eq:utility_decomposition}).
	
	\textit{Coverage and relative competence stability.}
	Theorem~1 separates changes in the distribution of representations from changes in competence at a given representation. Its change-of-measure argument also admits an integrated form. Suppose \(P_r^Z\ll P_s^Z\), write \(w=dP_r^Z/dP_s^Z\), and define
	\[
	\kappa_Z=\mathbb E_{P_s^Z}[w(Z)^2]<\infty,
	\qquad
	\Delta_{\mathrm{rel}}=\mathbb E_{P_r^Z}\operatorname{osc}(\boldsymbol\mu_r(Z)-\boldsymbol\mu_s(Z)).
	\]
	The first bound in Eq.(~\ref{eq:app_density_ratio}) with \(\mathbb E_{P_s^Z}[w^2]=\kappa_Z\) in place of \(\rho\), together with \(\operatorname{osc}(\boldsymbol\mu_r-\boldsymbol\mu_s)\) in place of \(2\delta\) in Eq.(~\ref{eq:app_domain_shift_regret}) and \(R_{r,k}\le\alpha_k\), yields
	\begin{align}
		\mathbb E_{P_r^Z}R_{r,k}(Z)
		&\le\min\!\left\{\alpha_k,\;C_{\tau,K,k}\sqrt{2\kappa_Z\mathcal E_s(\theta)}
		+\alpha_k(\beta_{\tau,r}+\Delta_{\mathrm{rel}})\right\},
		\label{eq:app_integrated_excess_bound}\\
		\mathbb E_{P_r^Z}R_{r,k}(Z)
		&\le\min\!\left\{\alpha_k,\;C_{\tau,K,k}\sqrt{2\kappa_Z\mathcal L_s(\theta)}
		+\alpha_k\Delta_{\mathrm{rel}}\right\}.
		\label{eq:app_integrated_total_bound}
	\end{align}
	The assumptions of Theorem~1 give \(\kappa_Z\le\rho\) and \(\Delta_{\mathrm{rel}}\le2\delta\). If \(\boldsymbol\mu_r(z)-\boldsymbol\mu_s(z)\) is constant across coordinates, then \(\Delta_{\mathrm{rel}}=0\) and the two Top-\(k\) sets of \(\boldsymbol\mu_r(z)\) and \(\boldsymbol\mu_s(z)\) coincide.
	
	\subsubsection{From Set Recovery to Score Fusion}
	\label{app:selection_fusion}
	
	Theorem~1 bounds the competence of the selected set. We next compare the resulting fused scores with those obtained from the conditionally competence-optimal set
	\(S_r^{*,(k)}(z)=\operatorname{TopK}(\boldsymbol\mu_r(z),k)\).
	Let \(\Psi\) be a measurable detection metric on the \([0,1]\) scale. For \(S\in\mathcal I_k\), define
	\begin{equation}
		F_{\mathcal T}(S)
		=\Psi\!\left(
		\mathcal M\!\left(\frac1k\sum_{j\in S}\mathcal N(\mathbf a_j(\mathbf X))\right),
		\mathbf y\right)\in[0,1],
		\label{eq:app_fusion_utility}
	\end{equation}
	where scores and labels are restricted to the evaluation interval, and \(\mathcal N\) and \(\mathcal M\) are the standardization and min-max transformations in Appendix~\ref{app:algorithm}. Under a fixed fitting and normalization protocol, identical selected sets yield identical fused scores.
	
	\textbf{Corollary 1 (Boundary Separation and Fused-Utility Agreement).}
	For \(1\le k<K\), write \(\mu_{r,(1)}(z)\ge\cdots\ge\mu_{r,(K)}(z)\) and define the boundary gap
	\begin{equation}
		\gamma_k(z)=\mu_{r,(k)}(z)-\mu_{r,(k+1)}(z).
		\label{eq:app_boundary_gap}
	\end{equation}
	Under the assumptions of Theorem~1, for every \(t>0\),
	\begin{align}
		&\mathbb E_{P_r}\left|
		F_{\mathcal T}(S_r^{*,(k)}(Z))
		-F_{\mathcal T}(\widehat S_\theta^{(k)}(Z))\right|\nonumber\\
		&\qquad\le
		\Pr_{P_r^Z}\!\left(\widehat S_\theta^{(k)}(Z)\ne S_r^{*,(k)}(Z)\right)
		\le\min\!\left\{1,\;\Pr_{P_r^Z}(\gamma_k(Z)\le t)
		+\frac{k\epsilon_k(\theta)}{t}\right\}.
		\label{eq:app_fusion_agreement}
	\end{align}
	For \(k=K\), the two sets and their fused outputs coincide identically.
	
	\textit{Proof.}
	On the event \(\gamma_k(z)>t\), the set \(S_r^{*,(k)}(z)\) is the unique collection of the \(k\) largest coordinates of \(\boldsymbol\mu_r(z)\), and every member has conditional competence more than \(t\) above every excluded specialist. If \(r\ge1\) members are replaced, pairing each missed index \(i\) with an incorrectly selected index \(\pi(i)\) gives
	\[
	\mu_{r,i}(z)-\mu_{r,\pi(i)}(z)\ge\gamma_k(z)>t,
	\]
	and therefore
	\[
	R_{r,k}(z)\ge\frac{r\gamma_k(z)}{k}>\frac{t}{k}.
	\]
	Splitting the mismatch event according to whether \(\gamma_k(z)\le t\) then yields
	\[
	\mathbf1\{\widehat S_\theta^{(k)}(z)\ne S_r^{*,(k)}(z)\}
	\le\mathbf1\{\gamma_k(z)\le t\}+\frac{kR_{r,k}(z)}{t}.
	\]
	Taking expectations and applying Theorem~1 bounds the mismatch probability by \(\Pr_{P_r^Z}(\gamma_k(Z)\le t)+k\epsilon_k(\theta)/t\). Since \(F_{\mathcal T}\in[0,1]\), its absolute difference is at most one on a mismatch and is zero when the sets agree, giving the first inequality. The probability is at most one, which proves Eq.(~\ref{eq:app_fusion_agreement}).
	\hfill\(\square\)
	
	The mismatch bound depends on the gap \(\gamma_k\) between the \(k\)-th and \((k+1)\)-th specialists. If \(\epsilon_k(\theta_n)\to0\), then for every \(t>0\),
	\[
	\limsup_{n\to\infty}
	\Pr_{P_r^Z}\!\bigl(\widehat S_{\theta_n}^{(k)}(Z)\ne S_r^{*,(k)}(Z)\bigr)
	\le\Pr_{P_r^Z}(\gamma_k(Z)\le t).
	\]
	Sending \(t\downarrow0\) and using \(\Pr_{P_r^Z}(\gamma_k(Z)=0)=0\) yields set recovery and convergence of the expected absolute fused-utility difference.
	
	\subsubsection{Scope of the Guarantee and Experimental Interpretation}
	\label{app:theory_scope}
	
	Mean individual competence \(V_k\) and the fused detection utility \(F_{\mathcal T}\) in Eq.(~\ref{eq:app_fusion_utility}) are different objectives. Theorem~1 compares selectors at the same budget through \(V_k\). Corollary~1 bounds the fused-score difference that arises when the selected set differs from \(S_r^{*,(k)}\); it does not rank detection performance across values of \(k\). The empirical budget sweep in Appendix~\ref{app:topk} evaluates that additional fusion tradeoff. The reported NDCG and Hit values assess predicted specialist rankings, while the detection metrics assess the complete fitting-and-fusion pipeline. The core analysis concerns a single routed series and applies to each variable-wise selection in the multivariate protocol.
	
	\clearpage
	\subsection{Overall Training and Deployment Procedure}
	\label{app:algorithm}
	
	Algorithm~\ref{alg:tsrouter} summarizes the complete competence-construction,
	router-training, and target-deployment procedure of TS-Router. The procedure
	assumes access to a pretrained temporal encoder \(E_{\mathrm{TSFM}}\), which is
	kept frozen throughout competence learning. In our main experiments,
	\(E_{\mathrm{TSFM}}\) is instantiated with the pretrained TimeRCD encoder.
	The labeled simulated tasks are used only to evaluate the relative competence
	of unsupervised specialists and construct routing targets; they do not provide
	supervision for fitting the specialists themselves. At deployment, routing is
	performed before detector fitting, so only the selected specialists are fitted
	on the training prefix of the unlabeled target series.
\begin{algorithm}[t]
    \caption{Training and deployment of TS-Router.}
    \label{alg:tsrouter}
    \begin{algorithmic}[1]

        \Require Simulated tasks
        \(\mathcal{D}_{\mathrm{sim}}
        =\{(\mathbf{X}_i^{\mathrm{sim}},
        \mathbf{y}_i^{\mathrm{sim}})\}_{i=1}^{N}\);
        pretrained encoder \(E_{\mathrm{TSFM}}\);
        specialist pool
        \(\mathcal{M}=\{M_1,\ldots,M_K\}\);
        competence utility \(\Phi\);
        temperature \(\tau\);
        number of selected specialists \(k\)

        \Ensure Trained router \(g_\theta\) and target anomaly scores

        \Statex
        \State \textbf{Stage I: Synthetic competence construction}
        \State Freeze \(E_{\mathrm{TSFM}}\)

        \For{each simulated task
            \((\mathbf{X}_i^{\mathrm{sim}},
            \mathbf{y}_i^{\mathrm{sim}})
            \in\mathcal{D}_{\mathrm{sim}}\)}

            \State Split \(\mathbf{X}_i^{\mathrm{sim}}\)
            into a fitting prefix \(\mathbf{X}_i^{\mathrm{sim,tr}}\)
            and an evaluation suffix \(\mathbf{X}_i^{\mathrm{sim,te}}\)
            without using anomaly labels

            \State Let \(\mathbf{y}_i^{\mathrm{sim,te}}\)
            denote the labels restricted to the evaluation suffix

            \For{each specialist \(M_j\in\mathcal{M}\)}
                \State Fit \(M_j\) to
                \(\mathbf{X}_i^{\mathrm{sim,tr}}\)
                without anomaly labels

                \State Obtain anomaly scores
                \(\mathbf{a}_{i,j}^{\mathrm{sim}}
                =M_j(\mathbf{X}_i^{\mathrm{sim,te}})\)

                \State Compute competence
                \(c_{i,j}^{\mathrm{sim}}
                =\Phi(
                \mathbf{a}_{i,j}^{\mathrm{sim}},
                \mathbf{y}_i^{\mathrm{sim,te}})\)
            \EndFor

            \State Construct the soft competence target
            \Statex \hspace{\algorithmicindent}
            \(\displaystyle
            q_{i,j}
            =
            \frac{\exp(c_{i,j}^{\mathrm{sim}}/\tau)}
            {\sum_{\ell=1}^{K}
            \exp(c_{i,\ell}^{\mathrm{sim}}/\tau)},
            \quad j=1,\ldots,K
            \)

            \State Extract final-layer token representations
            from the complete unlabeled series
            \(\mathbf{H}_i
            =E_{\mathrm{TSFM}}(\mathbf{X}_i^{\mathrm{sim}})\)

            \State Mean-pool the token representations
            \Statex \hspace{\algorithmicindent}
            \(\displaystyle
            \mathbf{z}_i
            =
            \frac{1}{m}
            \sum_{t=1}^{m}\mathbf{h}_{i,t}
            \)

        \EndFor

        \Statex
        \State \textbf{Stage II: Router learning}

        \While{the router has not converged}
            \State Predict competence distributions
            \(\hat{\mathbf q}_i
            =
            \operatorname{Softmax}
            (g_\theta(\mathbf z_i))\)

            \State Update \(\theta\) by minimizing
            \Statex \hspace{\algorithmicindent}
            \(\displaystyle
            \mathcal{L}_{\mathrm{route}}
            =
            \frac{1}{N}
            \sum_{i=1}^{N}
            D_{\mathrm{KL}}
            \left(
            \mathbf q_i
            \,\Vert\,
            \hat{\mathbf q}_i
            \right)
            \)
        \EndWhile

        \Statex
        \State \textbf{Stage III: Label-free target deployment}
        \State Receive an unlabeled target series \(\mathbf X\)

        \State Extract and mean-pool its temporal representation
        \Statex \hspace{\algorithmicindent}
        \(\displaystyle
        \mathbf z
        =
        \operatorname{MeanPool}
        \left(E_{\mathrm{TSFM}}(\mathbf X)\right)
        \)

        \State Predict relative specialist competence
        \(\hat{\mathbf q}(\mathbf X)
        =
        \operatorname{Softmax}
        (g_\theta(\mathbf z))\)

        \State Select specialists before fitting
        \Statex \hspace{\algorithmicindent}
        \(\displaystyle
        \mathcal S_\theta(\mathbf X)
        =
        \{M_j:
        j\in\operatorname{TopK}
        (\hat{\mathbf q}(\mathbf X),k)\}
        \)

        \State Split \(\mathbf X\) into a training prefix
        \(\mathbf X^{\mathrm{tr}}\) and a test suffix
        \(\mathbf X^{\mathrm{te}}\)

        \For{each \(M_j\in\mathcal S_\theta(\mathbf X)\)}
            \State Fit \(M_j\) to \(\mathbf X^{\mathrm{tr}}\)
            without anomaly labels

            \State Obtain anomaly scores
            \(\mathbf a_j(\mathbf X^{\mathrm{te}})\)

            \State Normalize scores as
            \(\tilde{\mathbf a}_j(\mathbf X^{\mathrm{te}})
            =
            \mathcal N(
            \mathbf a_j(\mathbf X^{\mathrm{te}}))\)
        \EndFor

        \State Fuse the selected specialists
        \Statex \hspace{\algorithmicindent}
        \(\displaystyle
        \bar{\mathbf a}(\mathbf X^{\mathrm{te}})
        =
        \frac{1}{k}
        \sum_{M_j\in\mathcal S_\theta(\mathbf X)}
        \tilde{\mathbf a}_j(\mathbf X^{\mathrm{te}})
        \)

        \State Min-max rescale the fused scores
        \Statex \hspace{\algorithmicindent}
        \(\displaystyle
        \hat{\mathbf a}(\mathbf X^{\mathrm{te}})
        =
        \mathcal M(
        \bar{\mathbf a}(\mathbf X^{\mathrm{te}}))
        \)

        \State \textbf{return}
        \(\hat{\mathbf a}(\mathbf X^{\mathrm{te}})\)

    \end{algorithmic}
\end{algorithm}
	
	For the main experiments, we set \(k=3\). Importantly, the complete specialist
	pool is evaluated when constructing synthetic competence supervision, whereas
	real-target deployment first predicts specialist suitability and then fits only
	the selected Top-\(k\) specialists. Thus, target anomaly labels and
	deployment-time evaluation of the complete candidate pool are both unnecessary
	for routing.
	
	\textbf{Score Normalization.}
	Because heterogeneous specialists may produce anomaly scores on different
	scales, each anomaly-score sequence is independently standardized before
	fusion. For a score sequence
	\(\mathbf a=(a_1,\ldots,a_L)\), we define
	\begin{equation}
		\mathcal N(\mathbf a)_t
		=
		\frac{a_t-\mu(\mathbf a)}
		{\sigma(\mathbf a)},
		\qquad t=1,\ldots,L,
		\label{eq:app_score_normalization}
	\end{equation}
	where \(\mu(\mathbf a)\) and \(\sigma(\mathbf a)\) denote the mean and standard
	deviation of the score sequence within the current segment. The standardized
	scores of the selected specialists are then combined by an unweighted mean.
	The fused sequence \(\bar{\mathbf a}\) is subsequently min-max rescaled to
	\([0,1]\):
	\begin{equation}
		\mathcal M(\bar{\mathbf a})_t
		=
		\frac{\bar a_t-\min(\bar{\mathbf a})}
		{\max(\bar{\mathbf a})-\min(\bar{\mathbf a})},
		\qquad t=1,\ldots,L,
		\label{eq:app_score_minmax}
	\end{equation}
	so that
	\begin{equation}
		\hat{\mathbf a}(\mathbf X)
		=
		\mathcal M
		\!\left(
		\frac{1}{k}
		\sum_{M_j\in\mathcal{S}_{\theta}(\mathbf X)}
		\mathcal N(\mathbf a_j(\mathbf X))
		\right).
		\label{eq:app_fused_score}
	\end{equation}
	For multivariate targets, the same per-specialist standardization is applied
	independently for each variable. Neither normalization nor fusion uses target
	anomaly labels.
	
	\textbf{Multivariate Targets.}
	For a multivariate target
	\(\mathbf{X}\in\mathbb{R}^{L\times V}\), we apply the same routing and
	detection procedure independently to each variable
	\(\mathbf{X}^{(v)}\), \(v=1,\ldots,V\). Each variable therefore has its own
	predicted competence profile and selected specialist subset,
	\begin{equation}
		\mathcal S_\theta(\mathbf X^{(v)})
		=\{M_j:j\in\operatorname{TopK}(\hat{\mathbf q}(\mathbf X^{(v)}),k)\}.
		\label{eq:app_variable_routing}
	\end{equation}
	The selected specialists are fitted independently to
	\(\mathbf{X}^{(v)}\). Their standardized scores are mean-fused and then
	min-max rescaled to produce the variable-wise score
	\begin{equation}
		\hat{\mathbf a}^{(v)}
		=
		\mathcal M
		\!\left(
		\frac{1}{k}
		\sum_{M_j\in
			\mathcal{S}_{\theta}(\mathbf{X}^{(v)})}
		\mathcal{N}
		\left(
		\mathbf a_j(\mathbf{X}^{(v)})
		\right)
		\right).
		\label{eq:app_variable_score}
	\end{equation}
	
	Before cross-variable aggregation, each variable-wise score sequence is again
	standardized with \(\mathcal N\). The standardized variable scores are
	averaged and then min-max rescaled:
	\begin{equation}
		\hat{\mathbf a}_{\mathrm{multi}}
		=
		\mathcal M
		\!\left(
		\frac{1}{V}
		\sum_{v=1}^{V}
		\mathcal{N}
		\left(
		\hat{\mathbf a}^{(v)}
		\right)
		\right).
		\label{eq:app_multivariate_aggregation}
	\end{equation}
	This two-stage normalization prevents differences in either specialist-score
	scale or variable-score scale from dominating the final prediction. The
	variable-wise treatment preserves the same routing mechanism across univariate
	and multivariate benchmarks without introducing a dimension-specific
	multivariate router.
	
	\clearpage
	\section{Additional Experimental Details}
	\label{app:experiments}
	
	\subsection{Experimental Setup}
	\label{app:experimental_setup}
	
	\subsubsection{Benchmark Datasets}
	\label{app:datasets}
	
	Our evaluation follows the TSB-AD benchmark and uses a collection of
	genuinely univariate and multivariate TSAD datasets. A substantial portion of
	the univariate track in TSB-AD is obtained by decomposing multivariate datasets
	into individual variables. To avoid evaluating closely related series in both
	the univariate and multivariate settings, we exclude these decomposed
	multivariate series from the univariate evaluation and retain only genuinely
	univariate collections. This results in eleven univariate benchmarks: IOPS,
	MGAB, NAB, NEK, Power, SED, Stock, TODS, UCR, WSD, and YAHOO. We additionally
	evaluate on five standard multivariate benchmarks: MSL, PSM, SMAP, SMD, and
	SWaT. Together, these datasets cover heterogeneous domains, temporal dynamics,
	and anomaly characteristics.
	
	The eleven genuinely univariate collections are also used for the controlled
	routing and representation analyses in the main text. This avoids introducing
	the additional variable-wise aggregation involved in multivariate detection
	when isolating the effects of specialist selection and representation choice.
	Tables~\ref{tab:univariate_datasets} and~\ref{tab:multivariate_datasets}
	summarize the dataset statistics, where \#TS denotes the number of time series,
	Avg. Length denotes the average number of observations per series, and AR
	denotes the anomaly ratio.
	
	\begin{table}[t]
		\caption{Statistics of the univariate benchmark datasets.}
		\label{tab:univariate_datasets}
		\centering
		\setlength{\tabcolsep}{4pt}
		\begin{tabular}{@{}llrrr@{}}
			\toprule
			Name & Domain & \#TS & Avg. Length & AR (\%) \\
			\midrule
			UCR   & Misc.       & 228 & 67818.7 & 0.6 \\
			NAB   & Mixed       & 28  & 5099.7  & 10.6 \\
			YAHOO & Web         & 259 & 1560.2  & 0.6 \\
			IOPS  & Operations  & 17  & 72792.3 & 1.3 \\
			MGAB  & Synthetic   & 9   & 97777.8 & 0.2 \\
			SED   & Medical     & 3   & 23332.3 & 4.1 \\
			Stock & Finance     & 20  & 15000.0 & 9.4 \\
			TODS  & Synthetic   & 15  & 5000.0  & 6.3 \\
			NEK   & Web         & 9   & 1073.0  & 8.0 \\
			Power & Power Grid  & 1   & 35040.0 & 8.5 \\
			WSD   & Web         & 111 & 17444.5 & 0.6 \\
			\bottomrule
		\end{tabular}
	\end{table}
	
	\begin{table}[t]
		\caption{Statistics of the multivariate benchmark datasets.}
		\label{tab:multivariate_datasets}
		\centering
		\setlength{\tabcolsep}{5pt}
		\begin{tabular}{@{}llrrr@{}}
			\toprule
			Name & Domain & \#TS & Avg. Length & AR (\%) \\
			\midrule
			MSL  & Space  & 16 & 3119.4   & 5.1  \\
			PSM  & Sensor & 1  & 217624.0 & 11.2 \\
			SMAP & Space  & 27 & 7855.9   & 2.9  \\
			SMD  & Server & 22 & 25466.4  & 3.8  \\
			SWaT & ICS    & 2  & 207457.5 & 12.7 \\
			\bottomrule
		\end{tabular}
	\end{table}
	
	\subsubsection{Baseline Methods and Deployment Protocols}
	\label{app:baselines}
	
	We compare TS-Router with two groups of TSAD baselines that represent different
	ways of using transferable or target-specific temporal knowledge. All compared
	methods are evaluated without access to target anomaly labels. However, their
	deployment protocols differ: direct zero-shot models apply a pretrained model
	without target-specific detector fitting, whereas target-fitted unsupervised
	methods learn or adapt their detection mechanism from each unlabeled target
	series. TS-Router belongs to neither category directly: it uses a fixed
	pretrained representation and routing policy to select specialists, after which
	only the selected specialists are fitted unsupervisedly to the target series.
	The comparison therefore evaluates different deployment paradigms under the
	common constraint of no target anomaly labels.
	
	\textbf{Direct Zero-Shot Models.}
	This group contains pretrained time-series models that can be applied to unseen
	target series without target-specific detector fitting.
	
	\begin{itemize}
		\item \textbf{TimeRCD} is an anomaly-specific time-series foundation model
		pretrained with labeled synthetic data for zero-shot anomaly detection.
		In the baseline setting, its pretrained detector is directly applied to
		each target without target-specific fitting.
		
		\item \textbf{DADA} is a pretrained general anomaly detector designed for
		cross-dataset deployment. Following the original experimental setting,
		we use an input window length of \(100\).
		
		\item \textbf{Chronos} formulates time-series forecasting through a
		language-modeling-style generative framework. We derive anomaly evidence
		from its pretrained predictive behavior and use an input window length
		of \(100\).
		
		\item \textbf{MOMENT} is a general-purpose time-series foundation model
		trained with patch-based temporal representation learning. We use an
		input window length of \(64\).
		
		\item \textbf{TimesFM} is a decoder-only pretrained time-series model
		designed primarily for zero-shot forecasting. We use an input window
		length of \(96\).
		
		\item \textbf{Time-MoE} is a decoder-only time-series foundation model
		with a sparse mixture-of-experts architecture. We use an input window
		length of \(96\).
	\end{itemize}
	
	\textbf{Target-Fitted Unsupervised Models.}
	This group contains representative deep TSAD methods that are fitted separately
	to each unlabeled target dataset or series. They therefore exploit
	target-specific normal temporal structure but do not use target anomaly labels.
	
	\begin{itemize}
		\item \textbf{OmniAnomaly} uses a stochastic recurrent architecture with
		variational latent variables to model normal multivariate temporal
		dynamics and detects anomalies through deviations from the learned
		normal model. We use an input window length of \(100\).
		
		\item \textbf{USAD} employs adversarially trained autoencoders for
		reconstruction-based anomaly detection. We use an input window length
		of \(100\).
		
		\item \textbf{TranAD} uses a transformer-based reconstruction framework
		to model temporal dependencies and scores anomalies from reconstruction
		discrepancies. We use an input window length of \(10\).
		
		\item \textbf{TFMAE} is a masked-autoencoder-based TSAD method that learns
		temporal representations by reconstructing masked time-series segments.
		
		\item \textbf{DCdetector} uses dual-attention contrastive representation
		learning to model temporal and channel-wise dependencies for anomaly
		detection.
	\end{itemize}
	
	For external baselines, we follow the configurations used in the original
	implementations or the benchmark protocol wherever applicable. Direct zero-shot
	models are not refitted on the target data, whereas target-fitted unsupervised
	models are trained independently on each target without anomaly labels. This
	distinction is preserved throughout all reported comparisons.
	
	\subsubsection{Specialist Pool}
	\label{app:specialists}
	
	TS-Router uses a fixed pool of eleven lightweight unsupervised anomaly
	detectors with heterogeneous inductive biases. Each detector is treated as a
	\emph{specialist} because it embodies a distinct criterion for modeling
	normality and measuring deviation, rather than corresponding to a predefined
	anomaly type. The pool spans subspace reconstruction, marginal density,
	neighborhood structure, one-class boundaries, isolation, prototype distance,
	and spectral or temporal structure. This diversity provides complementary
	views of abnormality across heterogeneous time series.
	
	The use of lightweight specialists is intentional. By limiting differences in
	model capacity, routing primarily reflects which inductive bias is appropriate
	for a target series rather than which candidate has the largest model capacity.
	Their low fitting cost also supports the route-before-fit deployment of
	TS-Router, where only the selected specialists are fitted to the unlabeled
	target series.
	
	Table~\ref{tab:specialist_pool} summarizes the specialist pool and its primary
	detection bias.
	
	\begin{table}[t]
		\caption{
			Specialists and their primary anomaly-detection inductive biases.
		}
		\label{tab:specialist_pool}
		\centering
		\setlength{\tabcolsep}{4pt}
		\begin{tabular}{@{}lp{0.64\linewidth}@{}}
			\toprule
			Specialist & Primary inductive bias \\
			\midrule
			PCA
			& Low-dimensional subspace reconstruction; anomalies produce large
			reconstruction residuals. \\
			
			Sub-PCA
			& Windowed PCA on subsequences; anomalies produce large
			reconstruction residuals relative to local temporal structure. \\
			
			Sub-HBOS
			& Marginal density modeling through histogram-based statistics;
			low-density observations receive larger anomaly scores. \\
			
			POLY
			& Smooth temporal structure modeled by polynomial fitting;
			deviations from the fitted temporal trend indicate anomalies. \\
			
			Sub-KNN
			& Neighborhood-distance structure; observations far from their
			nearest neighbors are considered anomalous. \\
			
			Sub-LOF
			& Relative local density; anomalies exhibit substantially lower
			local density than their surrounding neighborhoods. \\
			
			KMeansAD
			& Prototype-based structure; anomaly scores are determined by
			distance from learned cluster centers. \\
			
			OCSVM
			& One-class decision boundary; anomalies lie outside the region
			representing the dominant normal data distribution. \\
			
			Sub-OCSVM
			& One-class boundary modeling on windowed subsequences. \\
			
			Sub-IForest
			& Isolation structure; anomalous observations require fewer random
			partitions to isolate. \\
			
			SR
			& Spectral residual structure; anomalies correspond to salient
			deviations from regular spectral-temporal patterns. \\
			\bottomrule
		\end{tabular}
	\end{table}
	
	The same specialist pool is used when constructing synthetic competence
	supervision and when deploying TS-Router on real targets. During competence
	construction, all eleven specialists are fitted independently to each labeled
	simulated task without using anomaly labels for fitting, and their VUS-PR
	performance defines the competence profile. At deployment, the router predicts
	the relative suitability of the same specialists before fitting and selects
	only the Top-\(3\) candidates for target-specific unsupervised adaptation.
	Detailed detector hyperparameters are provided in
	Appendix~\ref{app:implementation}.
	
	\subsubsection{Evaluation Metrics}
	\label{app:metrics}
	
	We evaluate anomaly detection performance using four complementary metrics:
	VUS-PR, Affiliation-F1, F1$_{\mathrm{T}}$, and Standard-F1. Time-series
	anomalies often occur as contiguous intervals rather than isolated points, so
	relying on a single point-wise metric may not adequately reflect temporal
	overlap, event localization, and tolerance to boundary shifts. The four metrics
	therefore provide complementary point-wise, event-level, and range-aware views
	of detection quality.
	
	\textbf{Standard-F1.}
	Standard-F1 is the conventional point-wise F1 score computed from true
	positives, false positives, and false negatives. It measures the harmonic mean
	of point-wise precision and recall and provides a basic measure of anomaly
	localization accuracy.
	
	\textbf{F1$_{\mathrm{T}}$.}
	F1$_{\mathrm{T}}$ is a time-series range-wise F1 measure that accounts for
	temporal overlap and proximity between predicted and ground-truth anomaly
	intervals. Compared with purely point-wise evaluation, it better reflects cases
	where anomalous intervals are partially detected or predictions are slightly
	shifted relative to the annotated regions.
	
	\textbf{Affiliation-F1.}
	Affiliation-F1 evaluates predicted anomaly events through an
	affiliation-based matching mechanism that accounts for their temporal
	proximity to ground-truth events. It is less sensitive to small boundary
	misalignments than point-wise F1 and therefore provides an event-oriented view
	of detection quality.
	
	\textbf{Oracle-threshold evaluation.}
	Standard-F1, F1$_{\mathrm{T}}$, and Affiliation-F1 are computed from binary
	predicted labels obtained by thresholding continuous anomaly scores. The
	choice of threshold rule changes the meaning of the resulting F1 values: a
	threshold selected from evaluation labels is not interchangeable with a
	label-free, pre-specified, or train-time threshold. Throughout this paper, we
	apply a uniform \emph{oracle-threshold} protocol to all three metrics and all
	compared methods. After a method has produced evaluation scores, we search
	candidate thresholds on those scores and retain the threshold that maximizes
	the corresponding F1 metric against the evaluation labels. This search is
	performed independently for each method, each target series, and each F1
	metric. Evaluation labels are used only to choose the threshold after scores
	have been produced; they are not used for encoder training, router training,
	specialist fitting, or routing. The reported F1 numbers should therefore be
	read as oracle-threshold evaluation rather than as operational performance
	when labels are unavailable at threshold selection. This convention does not
	use labels for model training and does not invalidate the
	threshold-independent VUS-PR results, but it does limit the strength of
	deployment claims that would require choosing a threshold without evaluation
	labels.
	
	\textbf{VUS-PR.}
	VUS-PR is a threshold-independent, range-aware metric based on the
	precision--recall curve. It evaluates detection performance over different
	temporal tolerance ranges and integrates the resulting precision--recall
	behavior into a volume-under-the-surface score. This makes it suitable for
	settings where anomaly intervals may exhibit temporal lag or boundary
	uncertainty. We additionally use VUS-PR as the competence utility
	\(\Phi\) when constructing routing supervision, so the router is trained from
	a single consistent competence criterion while its downstream detection
	performance is evaluated under all four metrics.
	
	For routing analysis, we further report NDCG@$k$ and Hit@$k$, which compare
	the specialist ranking predicted by the router with the ground-truth competence
	ranking obtained from specialists' VUS-PR performance on each target task.
	NDCG@$k$ evaluates the quality of the predicted ordering while giving greater
	weight to highly ranked specialists. Hit@$k$ equals \(1\) when the predicted
	top-\(k\) set contains a specialist that attains the highest VUS-PR on the
	target task, and equals \(0\) otherwise. Higher values indicate stronger
	agreement with the VUS-PR-based ground-truth specialist ranking.
	
	For Best Fixed and Oracle, routing quality is evaluated with \(k=1\), whereas
	Top-3 selection strategies are evaluated with \(k=3\). Router Top-1 and
	TS-Router Top-3 use the same predicted ranking. Table~\ref{tab:routing_effectiveness}
	reports that ranking's NDCG@3 and Hit@3 once, in the TS-Router Top-3 row;
	the Router Top-1 row reports its detection performance only. Sharing a ranking
	does not imply equality of ranking metrics evaluated at different cutoffs.
	
	\subsubsection{Implementation Details}
\label{app:implementation}

This section records the implementation settings used throughout the main
experiments. Unless otherwise stated, TS-Router uses the same frozen encoder,
router, specialist pool, Top-\(3\) selection, and fusion rule.

\textbf{Encoder and Competence Supervision.}
We instantiate \(E_{\mathrm{TSFM}}\) with the publicly released TimeRCD
encoder and keep it frozen during competence learning and deployment.
Given a series, we z-score normalize the input, extract the final-layer token
representations, and mean-pool them to a \(256\)-dimensional vector
\(\mathbf z\). Routing supervision is constructed from \(850{,}000\) labeled
Syn-RCD tasks. Each simulated series is partitioned into a specialist-fitting
prefix and an evaluation suffix. Every specialist is fitted on the prefix
without anomaly labels and evaluated by VUS-PR on the suffix, following the
same prefix-fitting and suffix-scoring protocol as at deployment.
The resulting competence profile is converted into a soft target with
temperature \(\tau=0.1\) as in Eq.~(\ref{eq:soft_competence}).
The encoder receives the complete unlabeled simulated series, including
both portions, while anomaly labels are used only to evaluate specialist
competence and construct routing targets.

\textbf{Router Architecture and Training.}
The router \(g_\theta\) is an MLP with hidden widths
\(1024\), \(512\), and \(256\) and ReLU activations. It maps
\(\mathbf z\in\mathbb{R}^{256}\) to \(K\) specialist logits, which are
converted into \(\hat{\mathbf q}\) by a softmax. Only the router parameters
are optimized. We minimize the average Kullback--Leibler objective in
Eq.~(\ref{eq:routing_loss}) with Adam, learning rate \(10^{-4}\), weight
decay \(10^{-5}\), batch size \(512\), and \(30\) training epochs.
The random seed is \(42\).

\textbf{Specialist Hyperparameters.}
The eleven specialists use a fixed hyperparameter configuration across
synthetic competence construction and real-target deployment. Sliding-window
lengths are obtained from the TSB-AD period rank
(\(\mathrm{periodicity}\)). Table~\ref{tab:specialist_hyperparameters}
lists the settings used in all reported experiments.

\begin{table}[t]
    \caption{
        Specialist hyperparameters used for competence construction and
        target deployment.
    }
    \label{tab:specialist_hyperparameters}
    \centering
    \setlength{\tabcolsep}{4pt}
    \begin{tabular}{@{}lp{0.64\linewidth}@{}}
        \toprule
        Specialist & Hyperparameters \\
        \midrule
        PCA
        & Period rank \(1\); all principal components. \\
        Sub-PCA
        & Period rank \(1\); all principal components. \\
        Sub-HBOS
        & Period rank \(1\); \(10\) histogram bins. \\
        POLY
        & Period rank \(1\); polynomial degree \(4\). \\
        Sub-KNN
        & Period rank \(2\); \(50\) neighbors. \\
        Sub-LOF
        & Period rank \(2\); \(30\) neighbors. \\
        KMeansAD
        & Period rank \(2\); \(10\) clusters. \\
        OCSVM
        & Period rank \(1\); RBF kernel; \(\nu=0.5\). \\
        Sub-OCSVM
        & Period rank \(2\); RBF kernel; \(\nu=0.5\). \\
        Sub-IForest
        & Period rank \(1\); \(150\) trees. \\
        SR
        & Period rank \(1\). \\
        \bottomrule
    \end{tabular}
\end{table}

\textbf{Deployment Protocol.}
Following TSB-AD, each target series is split into a training prefix and a
test suffix. The series is z-score normalized before specialist fitting.
Under the offline observation protocol, the frozen encoder and router
receive the complete unlabeled target series, including both portions,
and select the Top-\(3\) specialists before fitting. Only these specialists
are then fitted on the training prefix without anomaly labels and produce
anomaly scores on the test suffix. Their scores are standardized,
mean-fused, and min-max rescaled as in Appendix~\ref{app:algorithm}.
All detection metrics are computed on the test suffix.
Multivariate targets use the same variable-wise routing and two-stage
aggregation described there. The encoder and router remain frozen
throughout target deployment; target observations are used for forward
inference and specialist fitting and scoring, without updating the
encoder or router parameters.

\textbf{Baseline Configurations.}
Direct zero-shot models use the input window lengths stated in
Appendix~\ref{app:baselines}. For TimeRCD we additionally use window length
\(5000\) and batch size \(64\). Target-fitted unsupervised models use the
same window lengths as in Appendix~\ref{app:baselines}, with learning rates
\(0.002\) for OmniAnomaly, \(0.001\) for USAD, and \(0.0001\) for TranAD.
Remaining baseline settings follow the original implementations or the
TSB-AD protocol.

\subsection{Extended Experiments}
\label{app:extended_experiments}

This section provides additional empirical results complementing the aggregate
comparisons in the main text. We first report complete per-dataset detection
results, followed by controlled analyses of the router representation,
specialist complementarity, Top-\(k\) selection, specialist-pool size, and
algorithm-selection transfer.
	
	\subsubsection{Per-Dataset Performance}
	\label{app:full_results}
	
	Table~\ref{tab:per_dataset_results} reports the complete per-dataset results
	corresponding to the aggregate comparison in Table~\ref{tab:main_results}.
	The results reveal substantial variation in the relative strengths of different
	methods across datasets and evaluation metrics. No competing detector
	consistently dominates across heterogeneous targets, including both pretrained
	zero-shot models and target-fitted unsupervised methods. This variation is
	consistent with the motivation of TS-Router: different temporal contexts can
	favor different anomaly-detection criteria.
	
	These per-dataset results complement the average-rank analysis in the main text.
	
	\begin{table*}[t]
		\caption{
			Per-dataset performance across 16 real-world TSAD benchmarks.
			Best results are in \textbf{bold} and second-best results are
			\underline{underlined}.
		}
		\label{tab:per_dataset_results}
		\centering
		\scriptsize
		\setlength{\tabcolsep}{1.6pt}
		\renewcommand{\arraystretch}{1.03}
		\resizebox{\textwidth}{!}{%
			\begin{tabular}{@{}llcccccccccccccccc@{}}
				\toprule
				\multirow{2}{*}{Metric}
				& \multirow{2}{*}{Model}
				& \multicolumn{11}{c}{Univariate datasets}
				& \multicolumn{5}{c}{Multivariate datasets} \\
				\cmidrule(lr){3-13}
				\cmidrule(lr){14-18}
				&
				& IOPS & MGAB & NAB & NEK & Power & SED & Stock & TODS & UCR & WSD & YAHOO
				& MSL & PSM & SMAP & SMD & SWaT \\
				\midrule
				
				\multirow{12}{*}{VUS-PR}
				& \textbf{TS-Router}
				& \textbf{41.49} & \textbf{43.09} & \underline{51.97} & \textbf{76.92}
				& \textbf{22.40} & \underline{65.42} & 73.18 & 71.65
				& \textbf{39.60} & \textbf{56.91} & 77.32
				& 20.86 & \textbf{19.55} & \textbf{38.12} & \textbf{52.24} & 42.75 \\
				& TimeRCD
				& 20.23 & \underline{1.05} & 24.32 & 27.88
				& \underline{21.25} & \textbf{80.75} & 77.28 & \textbf{93.46}
				& \underline{23.09} & 21.77 & \underline{84.41}
				& 20.45 & \underline{18.69} & 22.68 & 37.03 & 17.58 \\
				& DADA
				& 24.97 & 0.57 & 24.73 & 46.85
				& 10.61 & 6.42 & \textbf{99.51} & 64.83
				& 2.94 & 33.42 & 70.74
				& 12.74 & 17.17 & 20.02 & 25.98 & 21.13 \\
				& Chronos
				& 19.00 & 0.60 & 23.76 & 31.80
				& 10.95 & 8.65 & 97.49 & 70.66
				& 6.56 & 18.81 & 83.54
				& 8.25 & 14.61 & 5.18 & 10.22 & 16.44 \\
				& MOMENT
				& \underline{37.35} & 0.56 & 45.38 & 67.74
				& 10.50 & 4.31 & 76.97 & 56.45
				& 6.17 & \underline{55.26} & 30.81
				& 9.32 & 16.48 & 8.97 & 15.96 & 14.90 \\
				& TimesFM
				& 19.56 & 0.58 & 24.01 & 35.02
				& 10.44 & 6.13 & \underline{98.39} & \underline{72.89}
				& 6.03 & 21.57 & \textbf{86.78}
				& 11.84 & 14.76 & 16.95 & 13.02 & 19.43 \\
				& Time-MoE
				& 16.63 & 0.52 & 22.62 & 19.76
				& 9.34 & 10.87 & 74.78 & 48.78
				& 2.10 & 10.93 & 20.90
				& 7.82 & 15.68 & 4.98 & 11.12 & 16.20 \\
				& OmniAnomaly
				& 25.35 & 0.64 & 27.17 & \underline{74.51}
				& 14.32 & 6.20 & 91.29 & 45.55
				& 2.40 & 16.37 & 29.26
				& \textbf{31.57} & 18.58 & \underline{28.07} & \underline{37.44} & 42.97 \\
				& USAD
				& 16.58 & 0.75 & \textbf{55.03} & 58.53
				& 18.68 & 4.37 & 74.53 & 56.36
				& 8.85 & 10.00 & 14.15
				& \underline{29.95} & 17.59 & 26.37 & 34.53 & \underline{44.73} \\
				& TranAD
				& 21.61 & 0.64 & 24.82 & 61.63
				& 13.04 & 5.75 & 78.08 & 47.33
				& 2.25 & 12.20 & 25.78
				& 14.78 & 16.49 & 13.37 & 28.34 & \textbf{47.37} \\
				& TFMAE
				& 5.32 & 0.64 & 15.68 & 17.81
				& 11.90 & 9.55 & 73.54 & 48.79
				& 2.57 & 5.36 & 25.93
				& 8.25 & 14.22 & 5.76 & 4.77 & 15.38 \\
				& DCdetector
				& 5.83 & 0.59 & 16.60 & 14.03
				& 12.32 & 9.37 & 74.16 & 46.66
				& 1.53 & 3.23 & 10.17
				& 7.01 & 14.49 & 4.21 & 4.66 & 15.04 \\
				\midrule
				
				\multirow{12}{*}{Aff.-F1}
				& \textbf{TS-Router}
				& \underline{90.09} & \textbf{90.72} & \textbf{93.14} & 86.19
				& \textbf{88.88} & \underline{94.81} & 68.46 & 75.48
				& \textbf{90.43} & \textbf{97.16} & 93.80
				& \textbf{83.61} & 77.09 & \underline{88.25} & 91.55 & \textbf{80.25} \\
				& TimeRCD
				& 83.28 & \underline{70.69} & 82.48 & 79.73
				& \underline{85.51} & \textbf{96.87} & 71.84 & 86.37
				& \underline{84.63} & 90.33 & \textbf{96.65}
				& 81.16 & \textbf{81.61} & 87.73 & \textbf{92.58} & 71.55 \\
				& DADA
				& 89.37 & 67.66 & 86.56 & \textbf{95.40}
				& 69.79 & 65.18 & \textbf{98.77} & 76.89
				& 72.21 & 93.92 & 92.20
				& 76.57 & \underline{81.27} & 76.92 & 83.74 & \underline{76.18} \\
				& Chronos
				& \textbf{90.12} & 67.89 & 86.66 & \underline{93.63}
				& 69.72 & 67.89 & 96.85 & \textbf{91.96}
				& 74.35 & 90.98 & \underline{96.34}
				& 75.52 & 70.88 & 72.22 & 75.31 & 70.43 \\
				& MOMENT
				& 87.54 & 66.76 & 90.45 & 92.26
				& 75.97 & 59.13 & 45.26 & 59.76
				& 75.77 & \underline{95.39} & 79.99
				& 74.55 & 65.79 & 77.42 & 74.00 & 70.17 \\
				& TimesFM
				& 81.88 & 66.95 & 79.73 & 90.49
				& 69.88 & 67.14 & \underline{97.53} & \underline{89.08}
				& 70.03 & 78.97 & 91.28
				& 20.35 & 71.24 & 45.44 & 62.85 & 44.37 \\
				& Time-MoE
				& 76.34 & 67.23 & 80.51 & 80.50
				& 71.19 & 60.98 & 63.28 & 54.68
				& 73.56 & 80.25 & 69.70
				& 69.85 & 54.74 & 74.38 & 69.97 & 64.37 \\
				& OmniAnomaly
				& 80.32 & 67.35 & \underline{92.35} & 86.30
				& 78.16 & 61.26 & 75.24 & 50.73
				& 73.53 & 78.02 & 71.31
				& \underline{83.15} & 58.17 & \textbf{91.38} & 85.82 & 73.39 \\
				& USAD
				& 71.08 & 67.81 & 91.54 & 71.13
				& 76.48 & 55.60 & 35.92 & 47.90
				& 76.00 & 65.10 & 53.05
				& 81.86 & 57.86 & 87.25 & 85.09 & 75.06 \\
				& TranAD
				& 83.19 & 67.28 & 90.28 & 85.02
				& 71.56 & 61.03 & 57.94 & 52.76
				& 73.31 & 84.34 & 76.08
				& 79.91 & 73.83 & 87.39 & \underline{92.20} & 75.37 \\
				& TFMAE
				& 78.25 & 67.50 & 75.99 & 76.91
				& 70.30 & 68.17 & 56.39 & 62.83
				& 70.60 & 80.25 & 76.87
				& 75.70 & 70.07 & 75.36 & 70.85 & 75.72 \\
				& DCdetector
				& 71.83 & 67.91 & 72.21 & 62.31
				& 69.75 & 72.20 & 55.79 & 57.81
				& 70.18 & 72.79 & 67.77
				& 67.74 & 67.32 & 67.10 & 69.55 & 71.07 \\
				\midrule
				
				\multirow{12}{*}{F1$_{\mathrm{T}}$}
				& \textbf{TS-Router}
				& 48.63 & \textbf{36.62} & \underline{56.81} & \textbf{84.18}
				& \textbf{30.26} & \underline{58.43} & 17.26 & 34.38
				& \textbf{47.83} & \textbf{57.87} & 73.58
				& 43.30 & 27.90 & 43.61 & \underline{52.26} & 45.04 \\
				& TimeRCD
				& 28.44 & 1.81 & 38.85 & 35.87
				& \underline{28.47} & \textbf{69.43} & 31.73 & \textbf{65.89}
				& \underline{34.30} & 35.04 & \textbf{85.86}
				& 42.47 & \textbf{37.98} & 33.74 & \textbf{53.91} & 30.28 \\
				& DADA
				& 42.50 & 0.91 & 37.24 & 47.98
				& 19.80 & 9.56 & \textbf{95.49} & 35.18
				& 7.22 & \underline{48.46} & 79.52
				& 34.58 & \underline{31.84} & 30.42 & 40.80 & 35.13 \\
				& Chronos
				& 45.45 & 1.10 & 36.10 & 33.16
				& 19.90 & 13.18 & \underline{89.30} & \underline{53.90}
				& 10.88 & 39.82 & 79.00
				& 15.59 & 25.42 & 11.72 & 17.32 & 28.88 \\
				& MOMENT
				& 33.15 & 0.80 & 52.27 & 63.66
				& 19.91 & 9.54 & 18.04 & 17.47
				& 13.02 & 41.98 & 11.69
				& 25.97 & 27.77 & 17.93 & 28.68 & 28.76 \\
				& TimesFM
				& \underline{48.95} & 0.93 & 36.74 & 36.63
				& 19.80 & 9.58 & 88.94 & 51.13
				& 10.78 & 41.38 & \underline{83.46}
				& 7.83 & 25.42 & 11.64 & 18.65 & 21.39 \\
				& Time-MoE
				& 25.95 & 0.63 & 38.70 & 15.78
				& 19.85 & 17.73 & 34.13 & 20.91
				& 8.29 & 22.60 & 37.11
				& 23.92 & 26.82 & 14.22 & 19.90 & 30.11 \\
				& OmniAnomaly
				& \textbf{51.17} & 1.61 & 40.09 & \underline{82.20}
				& 23.48 & 9.68 & 36.22 & 14.33
				& 8.47 & 34.79 & 24.16
				& \textbf{49.36} & 30.42 & \textbf{46.63} & 51.84 & 46.64 \\
				& USAD
				& 20.99 & \underline{4.07} & \textbf{61.46} & 70.64
				& 28.23 & 9.54 & 16.86 & 20.85
				& 14.63 & 14.18 & 9.35
				& \underline{48.71} & 28.96 & \underline{43.94} & 50.41 & \textbf{50.41} \\
				& TranAD
				& 22.63 & 1.65 & 37.28 & 69.97
				& 22.36 & 9.57 & 16.73 & 13.51
				& 7.75 & 20.94 & 8.41
				& 39.42 & 25.49 & 29.12 & 37.98 & \underline{49.58} \\
				& TFMAE
				& 19.41 & 1.07 & 33.04 & 31.11
				& 20.18 & 11.82 & 22.15 & 16.48
				& 5.90 & 19.11 & 23.85
				& 25.28 & 25.36 & 19.39 & 10.13 & 28.46 \\
				& DCdetector
				& 6.61 & 1.32 & 32.72 & 29.21
				& 21.13 & 10.53 & 16.07 & 16.40
				& 6.62 & 7.32 & 6.81
				& 23.24 & 25.34 & 15.73 & 9.47 & 28.64 \\
				\midrule
				
				\multirow{12}{*}{Std.-F1}
				& \textbf{TS-Router}
				& \underline{42.44} & \textbf{36.56} & \underline{52.01} & \textbf{75.05}
				& \textbf{29.98} & \underline{58.71} & 18.83 & 40.41
				& \textbf{43.88} & \textbf{56.41} & 70.78
				& 28.79 & 26.97 & \underline{40.30} & \underline{53.20} & 51.08 \\
				& TimeRCD
				& 24.22 & 1.62 & 27.70 & 33.05
				& \underline{28.59} & \textbf{69.88} & 32.61 & \textbf{67.02}
				& \underline{28.13} & 31.96 & \textbf{87.02}
				& 30.66 & 26.00 & 30.48 & 44.89 & 28.73 \\
				& DADA
				& 32.76 & 0.80 & 26.91 & 48.24
				& 15.99 & 2.69 & \textbf{95.59} & 28.18
				& 3.36 & \underline{45.06} & 79.30
				& 22.13 & 24.07 & 26.75 & 34.98 & 34.78 \\
				& Chronos
				& 32.69 & 0.99 & 26.22 & 33.54
				& 17.47 & 8.74 & \underline{89.41} & \underline{40.52}
				& 8.21 & 34.58 & 78.89
				& 11.63 & 22.27 & 9.62 & 17.50 & 24.03 \\
				& MOMENT
				& 30.69 & 0.67 & 44.75 & 63.85
				& 16.39 & 3.36 & 19.38 & 14.64
				& 9.00 & 41.42 & 10.54
				& 14.43 & 23.83 & 12.92 & 29.78 & 21.30 \\
				& TimesFM
				& 34.28 & 0.83 & 26.46 & 38.15
				& 16.73 & 2.96 & 89.13 & 40.08
				& 7.86 & 38.50 & \underline{84.44}
				& 5.75 & 22.18 & 10.46 & 18.65 & 22.84 \\
				& Time-MoE
				& 26.52 & 0.45 & 26.20 & 11.47
				& 12.16 & 17.73 & 34.32 & 16.38
				& 4.09 & 20.09 & 27.50
				& 12.85 & 24.80 & 9.01 & 21.62 & 23.58 \\
				& OmniAnomaly
				& \textbf{47.05} & 1.44 & 28.81 & \underline{74.03}
				& 23.50 & 0.43 & 38.59 & 12.65
				& 5.11 & 29.57 & 21.40
				& \textbf{39.10} & \textbf{30.43} & \textbf{40.50} & \textbf{57.06} & 55.93 \\
				& USAD
				& 30.66 & \underline{3.89} & \textbf{56.15} & 62.91
				& 28.24 & 3.41 & 17.99 & 23.87
				& 10.74 & 13.20 & 7.21
				& \underline{38.71} & \underline{28.41} & 38.66 & 53.06 & \textbf{62.82} \\
				& TranAD
				& 34.85 & 1.46 & 27.33 & 60.36
				& 22.36 & 2.63 & 16.23 & 11.94
				& 4.40 & 20.23 & 5.70
				& 29.60 & 25.63 & 25.11 & 43.99 & \underline{61.86} \\
				& TFMAE
				& 9.48 & 0.97 & 23.78 & 19.74
				& 20.14 & 11.87 & 23.32 & 14.88
				& 2.83 & 15.53 & 20.50
				& 15.68 & 25.39 & 12.58 & 9.16 & 27.08 \\
				& DCdetector
				& 5.19 & 1.21 & 24.02 & 17.37
				& 21.10 & 10.54 & 16.97 & 17.85
				& 3.18 & 4.64 & 4.14
				& 14.08 & 25.33 & 10.67 & 8.99 & 27.02 \\
				
				\bottomrule
			\end{tabular}%
		}
	\end{table*}
	
	\subsubsection{Detailed Representation Ablation}
	\label{app:representation}
	
	We further examine how the representation supplied to the router affects
	specialist competence estimation. Table~\ref{tab:appendix_representation_univariate}
	reports the complete per-dataset results corresponding to the representation
	ablation in Table~\ref{tab:routing_effectiveness}. All variants use the same
	MLP router, eleven-specialist pool, competence supervision, Top-\(3\) selection,
	and normalized mean fusion; only the input representation is changed. This
	controlled setting isolates whether the representation itself provides useful
	information for predicting specialist suitability.
	
	TimeRCD achieves the best average performance on VUS-PR and Standard-F1.
	MOMENT is strongest on Affiliation-F1 and F1$_{\mathrm{T}}$, remaining close
	to TimeRCD on the latter. These differences are also broadly distributed
	across datasets rather than arising from a single benchmark. Handcrafted
	representations such as Catch22, TSFresh, and BasicStats can be
	strong on particular datasets but exhibit larger variation across temporal
	domains. Chronos and MOMENT remain weaker on VUS-PR and Standard-F1.
	
	Together with the NDCG and Hit values reported in
	Table~\ref{tab:routing_effectiveness}, these results suggest that the advantage
	of the pretrained temporal representation is not limited to downstream score
	fusion. It also yields a more informative description of the target series
	for predicting specialist competence. This is consistent with using the
	representation to decide which specialist criterion to trust, rather than as
	a universal anomaly score.
	
	\begin{table*}[t]
		\centering
		\caption{
			Representation ablation on eleven univariate TSB-AD benchmarks.
			All variants use the same MLP, 11-specialist pool, and Top-3 normalized
			mean fusion; only the router representation changes.
			TimeRCD remains frozen.
			Best results are in \textbf{bold} and second-best results are
			\underline{underlined}.
		}
		\label{tab:appendix_representation_univariate}
		\scriptsize
		\setlength{\tabcolsep}{2.0pt}
		\renewcommand{\arraystretch}{1.03}
		\resizebox{0.9\textwidth}{!}{%
			\begin{tabular}{@{}llcccccccccccc@{}}
				\toprule
				\multirow{2}{*}{Metric}
				& \multirow{2}{*}{Representation}
				& \multicolumn{11}{c}{Univariate datasets}
				& \multirow{2}{*}{Avg.} \\
				\cmidrule(lr){3-13}
				& & IOPS & MGAB & NAB & NEK
				& Power & SED & Stock & TODS
				& UCR & WSD & YAHOO & \\
				\midrule
				
				\multirow{7}{*}{VUS-PR}
				& TimeRCD
				& \underline{41.49} & \textbf{43.09} & \textbf{51.97} & \underline{76.92}
				& \textbf{22.40} & \underline{65.42} & 73.18 & 71.65
				& \textbf{39.60} & 56.91 & \underline{77.32}
				& \textbf{56.36} \\
				& Chronos
				& 38.63 & 33.44 & 49.34 & 50.10
				& 15.99 & \textbf{94.18} & 71.64 & 78.81
				& 33.17 & 53.83 & 66.87
				& 53.27 \\
				& MOMENT
				& 40.73 & 33.44 & 46.83 & 33.93
				& 20.56 & \textbf{94.18} & 87.24 & 76.35
				& \underline{38.09} & \underline{57.55} & 63.87
				& \underline{53.89} \\
				& Catch22
				& \textbf{44.58} & 9.46 & 42.75 & 38.31
				& \underline{22.27} & \textbf{94.18} & 76.57 & \textbf{84.05}
				& 34.11 & 48.86 & 54.39
				& 49.96 \\
				& TSFresh
				& 36.98 & 35.44 & 35.53 & 50.10
				& 17.54 & 37.17 & \underline{88.43} & 73.84
				& 33.18 & 51.50 & \textbf{79.40}
				& 49.01 \\
				& BasicStats
				& 40.65 & \underline{37.51} & 43.69 & 42.82
				& 20.93 & 6.52 & \textbf{90.20} & 72.17
				& 27.98 & 41.45 & 76.36
				& 45.48 \\
				& MiniRocket
				& 32.08 & 12.56 & \underline{50.28} & \textbf{77.04}
				& 14.72 & 61.63 & 78.24 & \underline{83.86}
				& 32.26 & \textbf{57.57} & 64.19
				& 51.31 \\
				
				\midrule
				\multirow{7}{*}{Aff.-F1}
				& TimeRCD
				& 90.09 & \textbf{90.72} & 93.14 & \textbf{86.19}
				& \textbf{88.88} & \underline{94.81} & 68.46 & 75.48
				& \underline{90.43} & 97.16 & \underline{93.80}
				& 88.11 \\
				& Chronos
				& 88.80 & 85.47 & \underline{93.38} & 81.72
				& 85.66 & \textbf{98.67} & 68.45 & \textbf{80.91}
				& 87.75 & 96.21 & 92.73
				& 87.25 \\
				& MOMENT
				& 91.69 & 85.47 & 93.07 & 76.63
				& \underline{87.16} & \textbf{98.67} & 81.16 & 80.00
				& \textbf{90.56} & \underline{97.26} & 91.79
				& \textbf{88.50} \\
				& Catch22
				& 89.75 & 74.09 & 91.02 & 80.67
				& 78.37 & \textbf{98.67} & 68.70 & 78.61
				& 88.65 & 93.45 & 89.16
				& 84.65 \\
				& TSFresh
				& \underline{92.18} & \underline{88.27} & 93.05 & \underline{84.74}
				& 82.58 & 84.07 & \textbf{84.72} & \underline{80.58}
				& 88.83 & 96.54 & \textbf{94.17}
				& \underline{88.16} \\
				& BasicStats
				& \textbf{93.03} & 87.51 & 92.57 & 70.62
				& 83.15 & 68.21 & \underline{83.69} & 75.94
				& 83.76 & 94.91 & \textbf{94.17}
				& 84.32 \\
				& MiniRocket
				& 88.08 & 77.33 & \textbf{93.93} & 81.91
				& 86.08 & 91.37 & 69.65 & 79.55
				& 88.60 & \textbf{97.33} & 91.19
				& 85.91 \\
				
				\midrule
				\multirow{7}{*}{F1$_{\mathrm{T}}$}
				& TimeRCD
				& 48.63 & \underline{36.62} & \textbf{56.81} & \textbf{84.18}
				& \textbf{30.26} & \underline{58.43} & 17.26 & 34.38
				& \textbf{47.83} & 57.87 & 73.58
				& \underline{49.62} \\
				& Chronos
				& 44.97 & \textbf{36.67} & 53.78 & 60.33
				& 22.77 & \textbf{75.27} & 17.46 & \textbf{43.79}
				& 39.91 & 57.75 & 52.47
				& 45.93 \\
				& MOMENT
				& \underline{49.51} & \textbf{36.67} & 52.20 & 49.99
				& \underline{28.94} & \textbf{75.27} & 56.67 & \underline{42.05}
				& \underline{46.09} & \textbf{60.13} & 49.59
				& \textbf{49.74} \\
				& Catch22
				& 49.37 & 26.12 & 50.55 & 49.94
				& 27.99 & \textbf{75.27} & 18.73 & 38.57
				& 40.73 & 51.01 & 36.82
				& 42.28 \\
				& TSFresh
				& 49.21 & 32.95 & 41.77 & 56.30
				& 24.50 & 29.90 & \underline{66.62} & 39.60
				& 42.03 & 56.68 & \underline{78.91}
				& 47.13 \\
				& BasicStats
				& \textbf{52.98} & 32.47 & 50.97 & 54.98
				& 25.48 & 9.79 & \textbf{67.71} & 30.19
				& 35.89 & 50.33 & \textbf{79.76}
				& 44.60 \\
				& MiniRocket
				& 45.30 & 33.56 & \underline{56.06} & \underline{80.46}
				& 22.37 & 48.85 & 20.12 & 38.74
				& 39.99 & \underline{58.78} & 46.73
				& 44.63 \\
				
				\midrule
				\multirow{7}{*}{Std.-F1}
				& TimeRCD
				& \underline{42.44} & \underline{36.56} & \textbf{52.01} & \underline{75.05}
				& \textbf{29.98} & \underline{58.71} & 18.83 & 40.41
				& \textbf{43.88} & 56.41 & 70.78
				& \textbf{47.73} \\
				& Chronos
				& 38.11 & \textbf{36.68} & 49.33 & 49.39
				& 22.63 & \textbf{75.69} & 19.12 & 37.97
				& 35.49 & 56.28 & 49.95
				& 42.79 \\
				& MOMENT
				& 40.24 & \textbf{36.68} & 46.04 & 36.81
				& \underline{28.61} & \textbf{75.69} & 57.75 & 41.62
				& \underline{41.74} & \textbf{57.93} & 46.98
				& \underline{46.37} \\
				& Catch22
				& \textbf{44.02} & 26.20 & 45.34 & 36.98
				& 28.03 & \textbf{75.69} & 20.50 & \underline{44.76}
				& 35.78 & 50.24 & 33.91
				& 40.13 \\
				& TSFresh
				& 38.96 & 33.05 & 36.14 & 44.56
				& 24.53 & 29.98 & \underline{67.34} & 29.32
				& 36.97 & 54.25 & \underline{76.34}
				& 42.86 \\
				& BasicStats
				& 41.30 & 32.46 & 43.93 & 46.40
				& 25.84 & 9.80 & \textbf{68.39} & 30.98
				& 29.89 & 48.22 & \textbf{77.21}
				& 41.31 \\
				& MiniRocket
				& 41.96 & 33.62 & \underline{50.05} & \textbf{75.49}
				& 22.15 & 49.14 & 22.09 & \textbf{46.45}
				& 36.78 & \underline{56.94} & 44.19
				& 43.53 \\
				
				\bottomrule
			\end{tabular}%
		}
	\end{table*}
	
	\subsubsection{Specialist Complementarity}
	\label{app:complementarity}
	
	A key premise of TS-Router is that heterogeneous anomaly detectors provide
	complementary inductive biases rather than forming a uniformly ordered set of
	strong and weak models. Table~\ref{tab:appendix_specialist_pool} reports the
	complete performance of all eleven specialists on the univariate benchmarks.
	The relative strengths of individual specialists vary substantially across
	datasets. For example, subspace-based methods are particularly effective on
	NEK, neighborhood-based specialists perform strongly on MGAB and SED,
	KMeansAD is competitive on Power, and SR is especially strong on Stock and
	YAHOO. No individual specialist consistently dominates across targets.
	
	This complementarity also appears at the aggregate level. The best fixed
	individual specialist achieves average scores of \(39.68\), \(84.34\),
	\(41.01\), and \(36.43\) under VUS-PR, Affiliation-F1, F1$_{\mathrm{T}}$,
	and Standard-F1, respectively, whereas TS-Router reaches \(56.36\),
	\(88.11\), \(49.62\), and \(47.73\). Thus, the advantage of TS-Router cannot
	be explained by repeatedly selecting a single globally strong detector.
	Instead, routing exploits changes in relative specialist competence across
	temporal contexts.
	
	We additionally report a hindsight Oracle that selects the best individual
	specialist separately for each target using its ground-truth detection
	performance. Together with Best Fixed, it illustrates the scope for
	individual-specialist adaptation rather than an upper bound on fused
	detection.
	
	\begin{table*}[t] 
		\centering
		\footnotesize
		\caption{ 
			Specialist complementarity on eleven univariate TSB-AD benchmarks. 
			TS-Router uses Top-3 normalized mean fusion, while Oracle denotes 
			hindsight per-target selection of the best individual specialist. 
			Oracle is an individual-specialist reference rather than an upper bound 
			on the fused TS-Router prediction. 
			Best and second-best non-oracle results are in \textbf{bold} and 
			\underline{underlined}, respectively. 
		} 
		\label{tab:appendix_specialist_pool} 
		\setlength{\tabcolsep}{2.0pt} 
		\renewcommand{\arraystretch}{1.03} 
		\resizebox{0.8\textwidth}{!}{%
			\begin{tabular}{@{}llcccccccccccc@{}} 
				\toprule
				\multirow{2}{*}{Metric}
				& \multirow{2}{*}{Method}
				& \multicolumn{11}{c}{Univariate datasets}
				& \multirow{2}{*}{Avg.} \\
				\cmidrule(lr){3-13}
				&
				& IOPS & MGAB & NAB & NEK & Power & SED
				& Stock & TODS & UCR & WSD & YAHOO & \\
				\midrule
				
				\multirow{13}{*}{VUS-PR}
				& \textbf{TS-Router}
				& \textbf{41.49}
				& \underline{43.09}
				& \textbf{51.97}
				& 76.92
				& 22.40
				& 65.42
				& 73.18
				& \underline{71.65}
				& \textbf{39.60}
				& \textbf{56.91}
				& \textbf{77.32}
				& \textbf{56.36} \\
				
				\cmidrule(lr){2-14}
				& PCA
				& 23.21 & 0.60 & 45.77 & \underline{78.45}
				& 10.49 & 3.77 & \underline{80.92} & 54.01
				& 13.40 & 16.95 & 21.15 & 31.70 \\
				
				& Sub-PCA
				& 20.35 & 0.66 & \underline{49.82} & \textbf{83.68}
				& 10.49 & 4.00 & 77.81 & 52.82
				& 13.53 & 14.09 & 19.41 & 31.51 \\
				
				& Sub-HBOS
				& 4.97 & 0.50 & 33.26 & 19.43
				& 15.69 & 63.09 & 64.60 & 64.54
				& 15.07 & 2.21 & 12.08 & 26.86 \\
				
				& POLY
				& 27.51 & 0.70 & 46.91 & 55.64
				& 9.30 & 7.88 & 79.01 & 57.39
				& 15.80 & 35.24 & 27.89 & 33.02 \\
				
				& Sub-KNN
				& 8.93 & 24.70 & 34.07 & 31.82
				& \underline{24.97} & \textbf{88.02} & 70.05 & 67.33
				& 34.03 & 10.27 & 28.83 & 38.46 \\
				
				& Sub-LOF
				& \underline{27.62} & \textbf{48.14} & 39.92 & 41.83
				& 17.46 & 24.70 & 74.86 & 53.07
				& \underline{35.25} & \underline{46.18} & 27.45
				& \underline{39.68} \\
				
				& KMeansAD
				& 6.71 & 0.97 & 34.96 & 27.06
				& \textbf{37.05} & \underline{84.19} & 71.34 & 67.12
				& 34.77 & 9.57 & 46.58 & 38.21 \\
				
				& OCSVM
				& 7.81 & 0.61 & 27.17 & 18.30
				& 20.20 & 9.81 & 68.43 & \textbf{76.08}
				& 9.79 & 3.74 & 41.75 & 25.79 \\
				
				& Sub-OCSVM
				& 7.11 & 0.93 & 30.43 & 25.79
				& 19.83 & 6.97 & 70.42 & 68.73
				& 9.87 & 3.46 & 24.03 & 24.32 \\
				
				& Sub-IForest
				& 14.38 & 0.56 & 45.13 & 68.04
				& 9.79 & 55.74 & 75.94 & 48.35
				& 8.12 & 6.39 & 15.64 & 31.64 \\
				
				& SR
				& 24.72 & 0.79 & 24.69 & 48.36
				& 12.39 & 7.80 & \textbf{99.65} & 61.66
				& 7.59 & 23.07 & \underline{71.51} & 34.75 \\
				
				\cmidrule(lr){2-14}
				& Oracle
				& 45.14 & 48.14 & 71.69 & 83.73
				& 37.05 & 88.85 & 99.65 & 80.44
				& 54.85 & 60.11 & 78.94 & 68.05 \\
				
				\midrule
				
				\multirow{13}{*}{Aff.-F1}
				& \textbf{TS-Router}
				& \underline{90.09}
				& \underline{90.72}
				& \textbf{93.14}
				& 86.19
				& \textbf{88.88}
				& 94.81
				& 68.46
				& \underline{75.48}
				& \textbf{90.43}
				& \textbf{97.16}
				& \textbf{93.80}
				& \textbf{88.11} \\
				
				\cmidrule(lr){2-14}
				& PCA
				& 78.75 & 67.32 & 92.46 & \underline{86.48}
				& 72.59 & 67.14 & 70.65 & 71.97
				& 80.41 & 78.76 & 74.97 & 76.50 \\
				
				& Sub-PCA
				& 78.78 & 67.42 & \underline{93.09} & \textbf{87.36}
				& 72.59 & 67.14 & \underline{71.30} & 71.61
				& 80.32 & 77.86 & 73.99 & 76.50 \\
				
				& Sub-HBOS
				& 68.78 & 67.17 & 84.88 & 70.59
				& 78.27 & 92.35 & 68.01 & 72.67
				& 81.25 & 70.65 & 72.28 & 75.17 \\
				
				& POLY
				& 80.95 & 68.86 & 88.90 & 82.73
				& 71.17 & 64.89 & 69.18 & 69.43
				& 82.68 & 83.74 & 79.27 & 76.53 \\
				
				& Sub-KNN
				& 70.48 & 79.28 & 82.95 & 68.42
				& 86.52 & \textbf{98.64} & 68.24 & 73.99
				& 87.52 & 74.67 & 85.89 & 79.69 \\
				
				& Sub-LOF
				& 85.13 & \textbf{94.92} & 91.33 & 79.52
				& 87.97 & 82.36 & 67.92 & 71.63
				& \underline{90.29} & \underline{95.71} & 80.91
				& \underline{84.34} \\
				
				& KMeansAD
				& 70.23 & 69.27 & 82.46 & 67.44
				& \underline{88.58} & \underline{97.32} & 67.97 & 72.86
				& 85.08 & 73.74 & 87.13 & 78.37 \\
				
				& OCSVM
				& 71.21 & 67.11 & 82.49 & 67.16
				& 76.81 & 72.12 & 68.40 & 75.45
				& 79.06 & 72.81 & 88.19 & 74.62 \\
				
				& Sub-OCSVM
				& 70.75 & 68.41 & 84.94 & 71.69
				& 85.47 & 70.54 & 68.62 & 72.88
				& 78.60 & 72.91 & 81.64 & 75.13 \\
				
				& Sub-IForest
				& 73.75 & 67.11 & 91.02 & 79.50
				& 85.92 & 91.05 & 69.31 & 68.76
				& 77.49 & 75.01 & 72.01 & 77.36 \\
				
				& SR
				& \textbf{92.04} & 69.68 & 86.80 & 85.34
				& 75.92 & 68.22 & \textbf{99.33} & \textbf{83.39}
				& 74.23 & 92.14 & \underline{91.61} & 83.52 \\
				
				\cmidrule(lr){2-14}
				& Oracle
				& 94.44 & 94.92 & 96.78 & 87.97
				& 88.58 & 98.64 & 99.33 & 85.01
				& 97.14 & 98.75 & 95.58 & 94.29 \\
				
				\midrule
				
				\multirow{13}{*}{F1$_{\mathrm{T}}$}
				& \textbf{TS-Router}
				& \textbf{48.63}
				& \underline{36.62}
				& \underline{56.81}
				& 84.18
				& \underline{30.26}
				& 58.43
				& 17.26
				& \underline{34.38}
				& \textbf{47.83}
				& \textbf{57.87}
				& \underline{73.58}
				& \textbf{49.62} \\
				
				\cmidrule(lr){2-14}
				& PCA
				& 31.83 & 0.99 & 54.09 & \underline{85.76}
				& 20.30 & 9.54 & 20.13 & 18.36
				& 19.38 & 24.41 & 12.93 & 27.07 \\
				
				& Sub-PCA
				& 27.70 & 2.51 & \textbf{57.88} & \textbf{88.26}
				& 20.30 & 9.55 & \underline{20.56} & 17.91
				& 19.66 & 21.69 & 12.34 & 27.12 \\
				
				& Sub-HBOS
				& 6.70 & 0.76 & 44.63 & 33.91
				& 24.23 & 49.93 & 16.30 & 21.94
				& 21.89 & 5.48 & 8.71 & 21.32 \\
				
				& POLY
				& 32.07 & 1.30 & 51.62 & 68.16
				& 16.95 & 14.82 & 19.71 & 23.36
				& 20.62 & 35.42 & 10.75 & 26.80 \\
				
				& Sub-KNN
				& 7.06 & 34.69 & 46.55 & 45.63
				& 28.20 & \underline{63.62} & 16.58 & 26.45
				& 40.91 & 11.72 & 13.50 & 30.45 \\
				
				& Sub-LOF
				& 33.36 & \textbf{39.61} & 46.75 & 54.87
				& 23.59 & 17.86 & 18.05 & 16.42
				& \underline{45.18} & \underline{47.80} & 13.06
				& 32.41 \\
				
				& KMeansAD
				& 7.73 & 8.29 & 47.88 & 49.35
				& \textbf{36.65} & \textbf{70.88} & 16.55 & 28.73
				& 41.14 & 12.52 & 36.25 & 32.36 \\
				
				& OCSVM
				& 8.57 & 1.22 & 39.51 & 35.56
				& 27.48 & 17.92 & 16.73 & 28.60
				& 15.07 & 7.69 & 16.90 & 19.57 \\
				
				& Sub-OCSVM
				& 6.52 & 7.33 & 41.83 & 38.92
				& 26.83 & 16.14 & 17.45 & 26.29
				& 14.85 & 7.11 & 11.65 & 19.54 \\
				
				& Sub-IForest
				& 15.79 & 0.84 & 50.73 & 75.16
				& 19.79 & 44.83 & 18.19 & 14.28
				& 16.26 & 15.51 & 12.23 & 25.78 \\
				
				& SR
				& \underline{47.06} & 1.42 & 38.12 & 54.60
				& 20.67 & 9.59 & \textbf{97.94} & \textbf{45.90}
				& 13.79 & 40.84 & \textbf{81.17}
				& \underline{41.01} \\
				
				\cmidrule(lr){2-14}
				& Oracle
				& 57.01 & 39.81 & 75.53 & 88.37
				& 36.65 & 70.88 & 97.94 & 56.57
				& 63.52 & 62.87 & 83.29 & 66.59 \\
				
				\midrule
				
				\multirow{13}{*}{Std.-F1}
				& \textbf{TS-Router}
				& \textbf{42.44}
				& \underline{36.56}
				& \underline{52.01}
				& 75.05
				& \underline{29.98}
				& 58.71
				& 18.83
				& 40.41
				& \textbf{43.88}
				& \textbf{56.41}
				& \underline{70.78}
				& \textbf{47.73} \\
				
				\cmidrule(lr){2-14}
				& PCA
				& 32.98 & 0.86 & 49.33 & \underline{75.76}
				& 20.30 & 9.43 & 22.06 & 19.74
				& 15.61 & 25.18 & 10.50 & 25.61 \\
				
				& Sub-PCA
				& 29.43 & 2.39 & \textbf{53.48} & \textbf{80.30}
				& 20.30 & 9.50 & \underline{22.52} & 19.38
				& 15.79 & 21.48 & 9.93 & 25.86 \\
				
				& Sub-HBOS
				& 5.73 & 0.67 & 37.70 & 23.45
				& 24.16 & 50.08 & 17.72 & 30.86
				& 17.05 & 3.80 & 5.43 & 19.70 \\
				
				& POLY
				& 28.95 & 1.14 & 50.18 & 58.50
				& 16.95 & 14.82 & 21.65 & 29.39
				& 17.87 & 36.43 & 8.87 & 25.89 \\
				
				& Sub-KNN
				& 7.26 & 34.73 & 40.68 & 35.78
				& \underline{28.22} & \underline{63.81} & 18.11 & \underline{43.74}
				& 36.49 & 7.98 & 10.92 & 29.79 \\
				
				& Sub-LOF
				& 27.55 & \textbf{39.60} & 40.91 & 43.13
				& 23.46 & 17.87 & 19.81 & 18.98
				& \underline{40.48} & \underline{45.26} & 10.93
				& 29.82 \\
				
				& KMeansAD
				& 8.99 & 8.22 & 42.37 & 36.16
				& \textbf{36.65} & \textbf{70.93} & 17.93 & \textbf{44.40}
				& 36.97 & 9.63 & 34.10 & 31.49 \\
				
				& OCSVM
				& 8.16 & 1.11 & 31.91 & 23.81
				& 27.51 & 17.92 & 17.96 & 41.41
				& 11.67 & 4.80 & 14.22 & 18.23 \\
				
				& Sub-OCSVM
				& 5.78 & 7.24 & 34.61 & 26.83
				& 26.85 & 16.15 & 18.61 & 41.28
				& 11.85 & 4.37 & 8.88 & 18.40 \\
				
				& Sub-IForest
				& 25.02 & 0.73 & 45.35 & 64.21
				& 19.82 & 45.25 & 19.66 & 14.43
				& 11.56 & 11.75 & 9.05 & 24.26 \\
				
				& SR
				& \underline{33.82} & 1.28 & 29.96 & 47.42
				& 21.19 & 9.57 & \textbf{97.92} & 33.64
				& 10.28 & 37.29 & \textbf{78.41}
				& \underline{36.43} \\
				
				\cmidrule(lr){2-14}
				& Oracle
				& 51.66 & 39.82 & 72.51 & 80.40
				& 36.65 & 70.93 & 97.92 & 58.17
				& 59.26 & 62.24 & 80.97 & 64.59 \\
				
				\bottomrule
			\end{tabular}%
		}
	\end{table*}
	
	\subsubsection{Effect of Top-\texorpdfstring{$k$}{k} Selection}
	\label{app:topk}
	
\begin{figure*}[t]
		\centering
		\includegraphics[width=\textwidth]{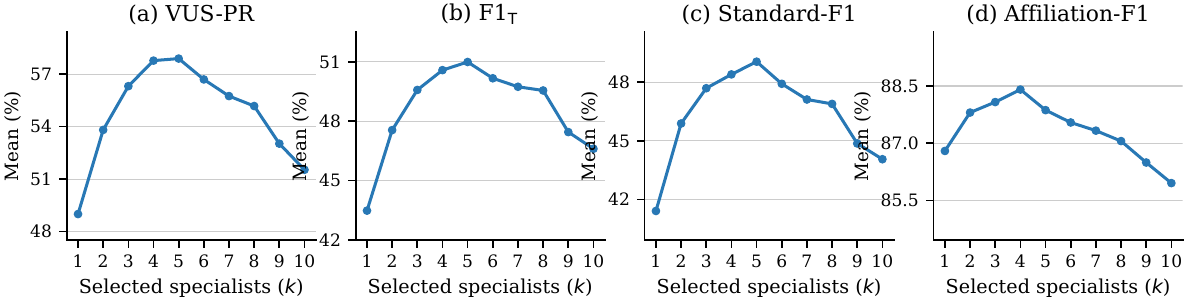}
		\caption{Effect of the number of selected specialists \(k\) with the
			11-specialist pool. Each panel reports the unweighted mean over eleven
			univariate benchmarks; the panels show VUS-PR, F1$_{\mathrm{T}}$,
			Standard-F1, and Affiliation-F1.}
		\label{fig:appendix_topk}
	\end{figure*}

	We sweep the number of selected specialists \(k\) from 1 to 10 while keeping
	the 11-specialist pool and the VUS-PR routing head fixed.
	Figure~\ref{fig:appendix_topk} reports dataset-averaged scores over the eleven
	univariate benchmarks.
	VUS-PR increases from \(49.00\) at \(k=1\); after a small selection budget,
	the scores decline as \(k\) grows toward \(10\). This sweep evaluates the
	empirical budget tradeoff in the complete selection-and-fusion pipeline.
	Theorem~1 compares set competence at a fixed \(k\), rather than predicting
	monotonic improvement of fused detection scores as the budget increases.
	
	\subsubsection{Effect of Specialist-Pool Size}
	\label{app:pool_size}
	
\begin{figure*}[t]
		\centering
		\includegraphics[width=\textwidth]{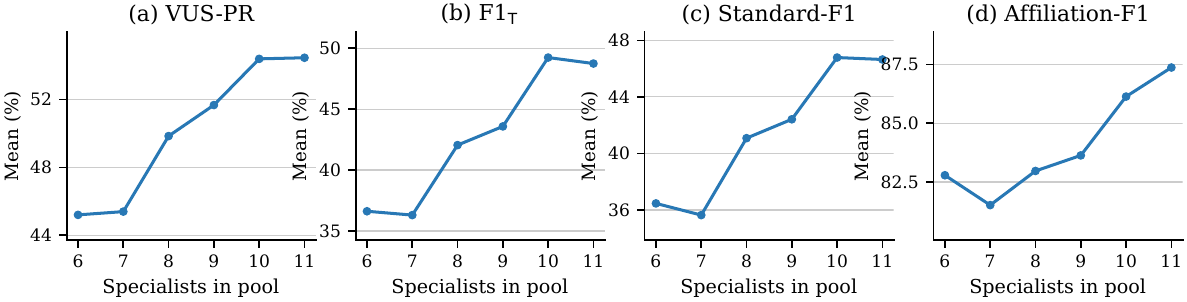}
		\caption{Effect of specialist-pool size with Top-3 selection. Each panel
			reports the unweighted mean over eleven univariate benchmarks; the panels
			show VUS-PR, F1$_{\mathrm{T}}$, Standard-F1, and Affiliation-F1. The router
			is retrained for each pool size.}
		\label{fig:appendix_pool_size}
	\end{figure*}

	We next vary the number of available specialists from 6 to 11. For each pool
	size, we retrain the router on the corresponding specialist pool and evaluate
	with Top-3 selection and the VUS-PR routing head.
	Figure~\ref{fig:appendix_pool_size} shows that enlarging the pool improves
	detection.
	
	\subsubsection{Router-Head Ablation}
	\label{app:head_ablation}
	
	We evaluate four routers that differ only in the metric used to construct the
	competence target: VUS-PR, Affiliation-F1, F1$_{\mathrm{T}}$ and Standard-F1.
	The specialist pool, frozen TimeRCD representation,
	Top-3 selection, and test set remain fixed. Each setting
	is evaluated on the same eleven univariate benchmarks as
	Table~\ref{tab:routing_effectiveness}. Table~\ref{tab:appendix_head_ablation}
	reports the downstream detection metrics, including the metric used as the
	routing head.
	
	The F1$_{\mathrm{T}}$ head gives the highest mean score under all five reported
	detection metrics, including VUS-PR (58.24) and Affiliation-F1 (89.17).
	The VUS-PR head remains close on VUS-PR (56.36), while the Standard-F1 head
	has the lowest VUS-PR, VUS-ROC and Affiliation-F1 values in this comparison.
	These results show that the supervision head changes the selected
	specialist rankings and can affect metrics beyond the head's own target.
	
	\begin{table}[t]
		\centering
		\small
		\caption{Top-3 routing with different competence-supervision heads. Values
			are dataset-averaged percentages over eleven univariate benchmarks. The pool
			size is 11 and the router is trained with the indicated head.}
		\label{tab:appendix_head_ablation}
		\setlength{\tabcolsep}{3.0pt}
		\begin{tabular}{@{}lccccc@{}}
			\toprule
			Routing head & VUS-PR & VUS-ROC & F1$_{\mathrm{T}}$ & Std.-F1 & Aff.-F1 \\
			\midrule
			VUS-PR         & 56.36 & 88.30 & 49.62 & 47.73 & 88.11 \\
			Affiliation-F1 & 56.27 & 89.24 & 49.17 & 46.83 & 88.49 \\
			F1$_{\mathrm{T}}$ & \textbf{58.24} & \textbf{89.29} & \textbf{53.03} & \textbf{49.96} & \textbf{89.17} \\
			Standard-F1   & 55.28 & 87.81 & 50.00 & 47.77 & 87.83 \\
			\bottomrule
		\end{tabular}
	\end{table}
	
	\subsubsection{Additional Algorithm-Selection Transfer Settings}
	\label{app:selector_transfer}
	
	We further compare TS-Router with classical algorithm-selection approaches
	under three complementary transfer settings. These experiments examine whether
	specialist competence can be transferred when selectors are provided with
	different forms of historical performance information. The three settings are
	not intended to impose identical supervision budgets. Instead, they
	progressively test selection under synthetic competence histories, limited
	labeled support from the target distribution, and labeled histories from other
	real datasets. In all settings, TS-Router keeps the same routing policy learned
	from Syn-RCD and receives no additional adaptation from the real support or
	target sets.

	\textbf{Synthetic-to-Real Selection.}
	We first consider the setting most directly aligned with the training protocol
	of TS-Router. Classical algorithm selectors are trained on Syn-RCD using
	specialist competence computed from the same eleven-specialist pool, and are
	then transferred directly to the eleven real univariate benchmarks.
	Neither approach receives real target anomaly labels for routing
	adaptation.
	
	Table~\ref{tab:appendix_selector_univariate} reports this comparison under a
	shared Top-\(3\) fusion protocol. TS-Router still achieves the
	highest average performance under all four detection metrics. Classical
	selectors remain competitive on particular datasets, indicating that
	synthetic competence histories contain useful transferable information, but
	their relative performance varies substantially across targets.
	
	\begin{table*}[t]
		\centering
		\caption{
			Algorithm-selection transfer on eleven univariate TSB-AD benchmarks.
			Classical selectors are trained on Syn-RCD using the common
			11-specialist pool. TS-Router uses its fixed routing policy
			learned from the same synthetic competence supervision.
			All selectors use Top-\(3\) fusion of the predicted specialists.
			Best results are in \textbf{bold} and second-best results are
			\underline{underlined}.
		}
		\label{tab:appendix_selector_univariate}
		\setlength{\tabcolsep}{2.2pt}
		\renewcommand{\arraystretch}{1.03}
		\resizebox{0.9\textwidth}{!}{%
			\begin{tabular}{@{}llcccccccccccc@{}}
				\toprule
				\multirow{2}{*}{Metric}
				& \multirow{2}{*}{Selector}
				& \multicolumn{11}{c}{Univariate datasets}
				& \multirow{2}{*}{Avg.} \\
				\cmidrule(lr){3-13}
				&
				& IOPS & MGAB & NAB & NEK & Power & SED & Stock & TODS
				& UCR & WSD & YAHOO & \\
				\midrule
				
				\multirow{7}{*}{VUS-PR}
				& \textbf{TS-Router}
				& 41.49 & \textbf{43.09} & \textbf{51.97} & \textbf{76.92}
				& \textbf{22.40} & 65.42 & 73.18 & 71.65
				& 39.60 & 56.91 & \textbf{77.32} & \textbf{56.36} \\
				& SATzilla
				& 41.77 & \underline{33.44} & 41.71 & 53.12
				& \textbf{22.40} & \textbf{74.65} & 70.43 & \textbf{72.73}
				& \underline{41.82} & \textbf{58.30} & \underline{59.20} & \underline{51.78} \\
				& ARGOSMART
				& \textbf{44.02} & 9.46 & \underline{44.73} & 31.94
				& \underline{22.36} & 11.94 & \textbf{82.21} & 70.71
				& 34.24 & 54.88 & 53.37 & 41.81 \\
				& MetaOD
				& 42.21 & \underline{33.44} & 40.23 & \underline{57.82}
				& \textbf{22.40} & 33.96 & 70.20 & \textbf{72.73}
				& \underline{41.82} & 56.72 & 57.89 & 48.13 \\
				& ISAC
				& \underline{43.02} & \underline{33.44} & 41.84 & 39.78
				& \textbf{22.40} & \textbf{74.65} & \underline{76.03} & \underline{72.72}
				& 40.42 & \underline{58.08} & 57.09 & 50.86 \\
				& MSAD
				& 36.70 & \underline{33.44} & 42.09 & 42.31
				& \textbf{22.40} & \underline{65.87} & 70.74 & 72.42
				& 40.98 & 52.49 & 56.72 & 48.74 \\
				& UReg
				& 42.12 & \underline{33.44} & 41.57 & 40.33
				& \textbf{22.40} & \textbf{74.65} & 70.48 & \textbf{72.73}
				& \textbf{41.83} & 51.18 & 53.89 & 49.51 \\
				\midrule
				\multirow{7}{*}{Aff.-F1}
				& \textbf{TS-Router}
				& 90.09 & \textbf{90.72} & \textbf{93.14} & \textbf{86.19}
				& \textbf{88.88} & 94.81 & \underline{68.46} & 75.48
				& 90.43 & 97.16 & \textbf{93.80} & \textbf{88.11} \\
				& SATzilla
				& 90.42 & \underline{85.47} & 91.38 & 81.61
				& \textbf{88.88} & \textbf{96.78} & 67.99 & \textbf{77.48}
				& \underline{92.06} & \textbf{97.58} & \underline{90.96} & \underline{87.33} \\
				& ARGOSMART
				& \underline{92.45} & 74.09 & \underline{92.75} & 75.75
				& \underline{84.13} & 71.40 & \textbf{73.97} & 75.77
				& 89.40 & 96.41 & 87.86 & 83.09 \\
				& MetaOD
				& 90.46 & \underline{85.47} & 91.59 & \underline{81.97}
				& \textbf{88.88} & 85.87 & 67.99 & \textbf{77.48}
				& \underline{92.06} & 97.11 & 90.70 & 86.32 \\
				& ISAC
				& \textbf{93.20} & \underline{85.47} & 92.15 & 76.88
				& \textbf{88.88} & \textbf{96.78} & 68.25 & \underline{77.19}
				& 91.80 & \underline{97.44} & 89.19 & 87.02 \\
				& MSAD
				& 88.48 & \underline{85.47} & 92.31 & 77.11
				& \textbf{88.88} & \underline{95.03} & 67.99 & 77.13
				& 91.82 & 95.77 & 90.54 & 86.41 \\
				& UReg
				& 86.39 & \underline{85.47} & 91.76 & 77.10
				& \textbf{88.88} & \textbf{96.78} & 67.99 & \textbf{77.48}
				& \textbf{92.11} & 94.75 & 89.71 & 86.22 \\
				\midrule
				\multirow{7}{*}{F1$_{\mathrm{T}}$}
				& \textbf{TS-Router}
				& 48.63 & \underline{36.62} & \textbf{56.81} & \textbf{84.18}
				& \textbf{30.26} & 58.43 & 17.26 & \textbf{34.38}
				& 47.83 & 57.87 & \textbf{73.58} & \textbf{49.62} \\
				& SATzilla
				& 49.56 & \textbf{36.67} & 48.45 & 62.45
				& \textbf{30.26} & \textbf{68.11} & 16.55 & \underline{34.32}
				& \underline{49.33} & \textbf{60.88} & 44.54 & \underline{45.56} \\
				& ARGOSMART
				& 49.83 & 26.12 & \underline{50.63} & 47.34
				& \underline{27.12} & 15.80 & \textbf{35.48} & 30.10
				& 42.14 & 58.05 & 44.55 & 38.83 \\
				& MetaOD
				& \underline{50.20} & \textbf{36.67} & 47.34 & \underline{66.39}
				& \textbf{30.26} & 28.79 & 16.46 & \underline{34.32}
				& \underline{49.33} & 59.79 & 41.95 & 41.96 \\
				& ISAC
				& \textbf{50.63} & \textbf{36.67} & 47.21 & 54.62
				& \textbf{30.26} & \textbf{68.11} & \underline{18.66} & 34.27
				& 48.36 & \underline{59.83} & \underline{48.66} & 45.21 \\
				& MSAD
				& 42.76 & \textbf{36.67} & 46.34 & 56.05
				& \textbf{30.26} & \underline{58.95} & 16.49 & 33.53
				& 48.98 & 54.83 & 37.84 & 42.06 \\
				& UReg
				& 43.45 & \textbf{36.67} & 48.29 & 53.95
				& \textbf{30.26} & \textbf{68.11} & 16.46 & \underline{34.32}
				& \textbf{49.39} & 53.22 & 35.23 & 42.67 \\
				\midrule
				\multirow{7}{*}{Std.-F1}
				& \textbf{TS-Router}
				& 42.44 & \underline{36.56} & \textbf{52.01} & \textbf{75.05}
				& \textbf{29.98} & 58.71 & 18.83 & \textbf{40.41}
				& 43.88 & 56.41 & \textbf{70.78} & \textbf{47.73} \\
				& SATzilla
				& 41.94 & \textbf{36.68} & 43.11 & 52.50
				& \textbf{29.98} & \textbf{68.39} & 18.13 & \underline{36.75}
				& \underline{45.45} & \textbf{58.58} & 41.13 & \underline{42.97} \\
				& ARGOSMART
				& 41.00 & 26.20 & \underline{44.37} & 34.55
				& \underline{27.52} & 15.77 & \textbf{36.92} & 36.00
				& 38.30 & 55.77 & 39.97 & 36.03 \\
				& MetaOD
				& \underline{42.47} & \textbf{36.68} & 41.45 & \underline{56.43}
				& \textbf{29.98} & 28.88 & 18.03 & \underline{36.75}
				& \underline{45.45} & 57.50 & 38.57 & 39.29 \\
				& ISAC
				& 41.29 & \textbf{36.68} & 41.41 & 41.46
				& \textbf{29.98} & \textbf{68.39} & \underline{20.48} & 36.71
				& 44.25 & \underline{57.63} & \underline{44.10} & 42.04 \\
				& MSAD
				& 37.75 & \textbf{36.68} & 40.94 & 42.89
				& \textbf{29.98} & \underline{59.18} & 18.06 & 35.56
				& 44.39 & 52.42 & 34.82 & 39.33 \\
				& UReg
				& \textbf{43.35} & \textbf{36.68} & 42.81 & 40.80
				& \textbf{29.98} & \textbf{68.39} & 18.03 & \underline{36.75}
				& \textbf{45.49} & 51.58 & 32.19 & 40.55 \\
				\bottomrule
			\end{tabular}%
		}
	\end{table*}

	\textbf{In-Distribution Selection with Limited Support.}
	We next consider a substantially more favorable setting for classical
	algorithm selection, restricted to the official TSB-AD Tuning split.
	This split provides the in-distribution subset used to develop detectors
	and selectors; after excluding decomposed multivariate series, it contains
	thirty genuinely univariate series from ten collections (IOPS, MGAB, NAB,
	NEK, SED, Stock, TODS, UCR, WSD, and YAHOO). Power does not appear in the
	Tuning list and is therefore omitted. Classical selectors may use a
	labeled support set of approximately \(10\%\) of series, drawn over ten
	random trials, to infer which specialists tend to perform well within
	that target distribution. We report their test-set scores on the Tuning
	series. TS-Router receives no support-set adaptation and is evaluated
	with exactly the same fixed policy as in the main experiments.
	
	As shown in Table~\ref{tab:id_support_transfer}, access to in-distribution
	support substantially narrows the gap for classical selectors. MetaOD
	reaches an average VUS-PR of \(55.91\), compared with \(57.40\) for
	TS-Router. MSAD attains the strongest Affiliation-F1, F1$_{\mathrm{T}}$,
	and Standard-F1 averages (\(90.92\), \(49.76\), and \(48.86\)), slightly
	above TS-Router at \(88.89\), \(48.79\), and \(48.01\). This result is
	informative because the classical selectors are now given labeled
	target-distribution history that TS-Router does not use. Even under this
	stronger supervision condition, the fixed TS-Router policy remains
	competitive across all four metrics, suggesting that the pretrained
	competence representation captures information that otherwise requires
	explicit target-specific selection history.
	
	\begin{table*}[t]
		\caption{
			In-distribution algorithm selection on the official
			TSB-AD Tuning subset. Evaluation uses the thirty
			genuinely univariate Tuning series from ten collections;
			Power does not appear in the Tuning split. Classical
			selectors are adapted from a small labeled support set
			across ten random trials, whereas TS-Router uses a
			fixed routing policy without support-set adaptation.
			Best results are in \textbf{bold} and second-best results are
			\underline{underlined}.
		}
		\label{tab:id_support_transfer}
		\centering
		\setlength{\tabcolsep}{3.0pt}
		\renewcommand{\arraystretch}{1.04}
		\resizebox{\textwidth}{!}{%
			\begin{tabular}{@{}llccccccccccc@{}}
				\toprule
				\multirow{2}{*}{Metric}
				& \multirow{2}{*}{Method}
				& \multicolumn{10}{c}{Univariate datasets}
				& \multirow{2}{*}{Avg.} \\
				\cmidrule(lr){3-12}
				&
				& IOPS & MGAB & NAB & NEK & SED
				& Stock & TODS & UCR & WSD & YAHOO & \\
				\midrule
				\multirow{7}{*}{VUS-PR}
				& \textbf{TS-Router}
				& \textbf{28.72} & \underline{45.61} & \textbf{57.45} & \textbf{94.24} & 63.78
				& 52.84 & \textbf{87.54} & 29.67 & \textbf{45.05} & \textbf{69.09} & \textbf{57.40} \\
				& SATzilla
				& 20.45 & \textbf{48.52} & \underline{43.74} & \underline{76.40} & \underline{77.57}
				& 90.36 & 73.87 & \underline{38.32} & 35.01 & 45.81 & 55.00 \\
				& ARGOSMART
				& \underline{23.07} & \textbf{48.52} & 42.86 & 41.35 & \underline{77.57}
				& \underline{94.57} & \underline{78.03} & 25.53 & \underline{38.48} & 49.53 & 51.95 \\
				& MetaOD
				& 19.58 & \textbf{48.52} & 39.00 & \underline{76.40} & \underline{77.57}
				& 90.36 & 77.94 & \textbf{39.62} & 35.01 & \underline{55.13} & \underline{55.91} \\
				& ISAC
				& 18.96 & 42.10 & 42.93 & 36.70 & 36.09
				& 52.00 & 64.30 & 31.10 & 33.26 & 19.34 & 37.68 \\
				& MSAD
				& 20.06 & \textbf{48.52} & 34.97 & 70.09 & \underline{77.57}
				& \textbf{99.21} & 65.82 & 33.31 & 36.73 & 47.28 & 53.36 \\
				& UReg
				& 19.38 & \textbf{48.52} & 39.95 & 34.09 & \textbf{89.25}
				& 52.09 & 62.08 & 32.39 & 35.77 & 16.16 & 42.97 \\
				\midrule
				\multirow{7}{*}{Aff.-F1}
				& \textbf{TS-Router}
				& \underline{86.08} & 90.12 & \textbf{94.44} & \textbf{99.16} & 93.25
				& 68.30 & \underline{82.30} & 83.45 & \textbf{98.96} & \textbf{92.82} & 88.89 \\
				& SATzilla
				& 85.15 & \textbf{94.46} & \underline{89.02} & 91.91 & \underline{96.18}
				& 94.29 & 81.59 & \underline{88.08} & 91.18 & 89.00 & \underline{90.09} \\
				& ARGOSMART
				& \textbf{87.88} & \textbf{94.46} & 83.80 & 84.29 & \underline{96.18}
				& \underline{96.12} & 81.99 & 83.44 & \underline{96.21} & 90.91 & 89.53 \\
				& MetaOD
				& 83.23 & \textbf{94.46} & 87.40 & 91.91 & \underline{96.18}
				& 94.29 & \textbf{82.43} & \textbf{88.44} & 91.18 & \underline{91.05} & 90.06 \\
				& ISAC
				& 82.40 & \underline{90.92} & 86.43 & 78.10 & 82.92
				& 67.72 & 79.72 & 84.75 & 90.78 & 86.13 & 82.99 \\
				& MSAD
				& 86.06 & \textbf{94.46} & 87.60 & \underline{94.27} & \underline{96.18}
				& \textbf{99.46} & 80.11 & 87.22 & 95.84 & 88.00 & \textbf{90.92} \\
				& UReg
				& 80.82 & \textbf{94.46} & 88.55 & 77.81 & \textbf{98.60}
				& 67.52 & 79.69 & 86.16 & 95.14 & 83.36 & 85.21 \\
				\midrule
				\multirow{7}{*}{F1$_{\mathrm{T}}$}
				& \textbf{TS-Router}
				& \textbf{44.30} & 36.99 & \textbf{62.44} & \textbf{87.59} & 47.05
				& 15.87 & \underline{40.34} & 26.97 & \textbf{62.03} & \textbf{64.34} & 48.79 \\
				& SATzilla
				& 25.47 & \textbf{43.95} & 50.67 & \underline{73.75} & \underline{55.04}
				& 85.13 & 37.60 & \underline{38.34} & 41.80 & 37.58 & \underline{48.93} \\
				& ARGOSMART
				& 29.88 & \textbf{43.95} & 48.16 & 43.53 & \underline{55.04}
				& \underline{89.69} & \textbf{40.50} & 29.43 & \underline{53.87} & 34.76 & 46.88 \\
				& MetaOD
				& 17.37 & \textbf{43.95} & 46.84 & \underline{73.75} & \underline{55.04}
				& 85.13 & 40.26 & \textbf{38.84} & 41.80 & \underline{44.83} & 48.78 \\
				& ISAC
				& 23.95 & \underline{42.14} & \underline{52.81} & 37.89 & 32.67
				& 17.79 & 30.14 & 31.51 & 44.69 & 7.44 & 32.10 \\
				& MSAD
				& \underline{30.24} & \textbf{43.95} & 44.51 & 65.55 & \underline{55.04}
				& \textbf{98.66} & 32.26 & 35.71 & 51.47 & 40.26 & \textbf{49.76} \\
				& UReg
				& 17.26 & \textbf{43.95} & 47.95 & 34.77 & \textbf{65.87}
				& 17.60 & 27.63 & 34.00 & 52.48 & 9.14 & 35.06 \\
				\midrule
				\multirow{7}{*}{Std.-F1}
				& \textbf{TS-Router}
				& \textbf{43.85} & 37.09 & \textbf{57.00} & \textbf{86.73} & 47.35
				& 17.33 & \underline{40.32} & 24.66 & \textbf{62.06} & \textbf{63.74} & 48.01 \\
				& SATzilla
				& 25.65 & \textbf{43.95} & 45.23 & \underline{74.20} & \underline{55.31}
				& 85.34 & 37.48 & \underline{34.39} & 42.69 & 37.95 & \underline{48.22} \\
				& ARGOSMART
				& 30.18 & \textbf{43.95} & 46.15 & 44.48 & \underline{55.31}
				& \underline{89.80} & \textbf{40.71} & 25.48 & \underline{54.68} & 34.75 & 46.55 \\
				& MetaOD
				& 17.24 & \textbf{43.95} & 41.20 & \underline{74.20} & \underline{55.31}
				& 85.34 & 40.28 & \textbf{35.25} & 42.69 & \underline{45.39} & 48.08 \\
				& ISAC
				& 24.12 & \underline{42.19} & \underline{46.29} & 41.13 & 32.79
				& 19.52 & 30.01 & 26.71 & 45.81 & 7.36 & 31.59 \\
				& MSAD
				& \underline{30.56} & \textbf{43.95} & 38.00 & 65.42 & \underline{55.31}
				& \textbf{98.58} & 31.91 & 31.94 & 52.44 & 40.46 & \textbf{48.86} \\
				& UReg
				& 17.32 & \textbf{43.95} & 41.10 & 38.12 & \textbf{66.12}
				& 19.40 & 27.47 & 28.63 & 53.69 & 9.02 & 34.48 \\
				\bottomrule
			\end{tabular}%
		}
	\end{table*}

	\textbf{Leave-One-Dataset-Out Transfer.}
Finally, we examine cross-dataset transfer using real historical
specialist performance. For each of the eleven univariate benchmarks,
the complete dataset is held out as an unseen target, while classical
selectors are trained on specialist-performance records from the
remaining ten datasets. Each classical selector deploys its top-ranked
specialist (Top-1), while TS-Router uses its default Top-3 normalized
mean fusion. TS-Router retains its routing policy learned on Syn-RCD,
without real-data retraining or target-specific adaptation of the router.
This comparison evaluates the complete detection pipelines under
cross-dataset transfer.

Table~\ref{tab:ood_selection_transfer} shows that TS-Router achieves
the highest dataset-averaged performance across all four metrics.
The strongest classical-selector averages are \(38.43\), \(80.99\),
\(33.17\), and \(31.43\) under VUS-PR, Affiliation-F1,
F1$_{\mathrm{T}}$, and Standard-F1, respectively, compared with
\(56.36\), \(88.11\), \(49.62\), and \(47.73\) for TS-Router.

Moreover, the Router Top-1 variant reported in Table~2 also exceeds
the strongest classical-selector average under each of the four metrics,
showing that the advantage persists with single-specialist selection.
These findings support the cross-dataset transferability of a
competence-routing policy learned from pretrained temporal representations
and synthetic supervision, without retraining the router on real-world
detector-performance histories.

\begin{table*}[t]
    \caption{
        Leave-one-dataset-out transfer across eleven univariate benchmarks.
        Classical selectors are trained on the other ten real benchmark
        collections and deploy their top-ranked specialist (Top-1).
        TS-Router uses its fixed routing policy learned on Syn-RCD and
        default Top-3 normalized mean fusion, without real-data retraining
        of the router. All methods are evaluated on the entirely held-out
        target dataset.
        Best results are in \textbf{bold} and second-best results are
        \underline{underlined}.
    }
    \label{tab:ood_selection_transfer}
    \centering
    \setlength{\tabcolsep}{2.4pt}
    \renewcommand{\arraystretch}{1.03}
    \resizebox{\textwidth}{!}{%
        \begin{tabular}{@{}llcccccccccccc@{}}
            \toprule
            \multirow{2}{*}{Metric}
            & \multirow{2}{*}{Method}
            & \multicolumn{11}{c}{Held-out target dataset}
            & \multirow{2}{*}{Avg.} \\
            \cmidrule(lr){3-13}
            &
            & IOPS & MGAB & NAB & NEK & Power & SED
            & Stock & TODS & UCR & WSD & YAHOO & \\
            \midrule

            \multirow{7}{*}{VUS-PR}
            & \textbf{TS-Router}
            & \textbf{41.49} & \textbf{43.09} & \textbf{51.97}
            & \textbf{76.92} & 22.40 & \textbf{65.42}
            & 73.18 & \textbf{71.65} & \textbf{39.60}
            & \textbf{56.91} & \textbf{77.32}
            & \textbf{56.36} \\
            & SATzilla
            & 29.32 & 0.97 & 38.66
            & 39.31 & \underline{24.97} & 7.88
            & \underline{77.81} & 62.13 & 20.21
            & 43.48 & 29.34
            & 34.01 \\
            & ARGOSMART
            & 27.68 & \underline{24.70} & 32.40
            & \underline{52.31} & 9.30 & \underline{24.70}
            & 76.93 & 57.39 & 20.83
            & 44.15 & \underline{36.26}
            & 36.97 \\
            & MetaOD
            & 27.79 & 0.97 & 40.12
            & 39.31 & \underline{24.97} & 7.88
            & \underline{77.81} & 63.46 & 20.43
            & 44.46 & 29.37
            & 34.23 \\
            & ISAC
            & \underline{31.77} & \underline{24.70} & 41.41
            & 41.83 & \textbf{37.05} & 7.88
            & \textbf{78.19} & \underline{64.33} & \underline{25.04}
            & 40.10 & 30.47
            & \underline{38.43} \\
            & MSAD
            & 28.16 & 0.97 & 40.03
            & 40.37 & 9.30 & \underline{24.70}
            & 74.98 & 63.76 & 14.51
            & 46.07 & 29.46
            & 33.85 \\
            & UReg
            & 29.31 & 0.97 & \underline{42.83}
            & 41.83 & 17.46 & \underline{24.70}
            & 77.10 & 53.30 & 12.03
            & \underline{46.77} & 27.18
            & 33.95 \\
            \midrule

            \multirow{7}{*}{Aff.-F1}
            & \textbf{TS-Router}
            & \textbf{90.09} & \textbf{90.72} & \textbf{93.14}
            & \textbf{86.19} & \textbf{88.88} & \textbf{94.81}
            & 68.46 & \textbf{75.48} & \textbf{90.43}
            & \textbf{97.16} & \textbf{93.80}
            & \textbf{88.11} \\
            & SATzilla
            & 86.96 & 69.27 & 87.37
            & 77.77 & 86.52 & 64.89
            & \textbf{71.30} & 74.20 & 80.48
            & 93.63 & 81.50
            & 79.44 \\
            & ARGOSMART
            & 86.34 & \underline{79.28} & 84.17
            & 74.73 & 71.17 & \underline{82.36}
            & \underline{68.99} & 69.43 & 81.76
            & 94.43 & \underline{82.65}
            & 79.57 \\
            & MetaOD
            & 84.96 & 69.27 & 88.36
            & 77.77 & 86.52 & 64.89
            & \textbf{71.30} & 73.08 & 81.00
            & 94.04 & 81.59
            & 79.34 \\
            & ISAC
            & \underline{88.05} & \underline{79.28} & 91.35
            & \underline{79.52} & \underline{88.58} & 64.89
            & 68.93 & 73.40 & \underline{82.86}
            & 92.70 & 81.36
            & \underline{80.99} \\
            & MSAD
            & 86.49 & 69.27 & 90.61
            & 78.49 & 71.17 & \underline{82.36}
            & 67.73 & \underline{74.49} & 78.61
            & \underline{95.57} & 81.48
            & 79.66 \\
            & UReg
            & 85.62 & 69.27 & \underline{92.12}
            & \underline{79.52} & 87.97 & \underline{82.36}
            & 68.43 & 71.62 & 76.23
            & 95.45 & 80.92
            & 80.86 \\
            \midrule

            \multirow{7}{*}{F1$_{\mathrm{T}}$}
            & \textbf{TS-Router}
            & \textbf{48.63} & \textbf{36.62} & \textbf{56.81}
            & \textbf{84.18} & \underline{30.26} & \textbf{58.43}
            & 17.26 & \textbf{34.38} & \textbf{47.83}
            & \textbf{57.87} & \textbf{73.58}
            & \textbf{49.62} \\
            & SATzilla
            & 33.93 & 8.29 & \underline{50.00}
            & 58.02 & 28.20 & 14.82
            & \textbf{20.56} & \underline{28.38} & 25.49
            & 41.98 & 15.24
            & 29.54 \\
            & ARGOSMART
            & 33.01 & \underline{34.69} & 41.08
            & \underline{63.95} & 16.95 & \underline{17.86}
            & 18.75 & 23.36 & 27.31
            & 45.83 & \underline{23.26}
            & 31.46 \\
            & MetaOD
            & 34.68 & 8.29 & 49.05
            & 58.02 & 28.20 & 14.82
            & \textbf{20.56} & 25.04 & 25.96
            & 43.62 & 14.96
            & 29.38 \\
            & ISAC
            & \underline{39.96} & \underline{34.69} & 47.97
            & 54.87 & \textbf{36.65} & 14.82
            & \underline{19.07} & 25.75 & \underline{31.73}
            & 43.98 & 15.33
            & \underline{33.17} \\
            & MSAD
            & 34.15 & 8.29 & 47.71
            & 53.72 & 16.95 & \underline{17.86}
            & 18.01 & 27.43 & 20.69
            & 47.96 & 14.71
            & 27.95 \\
            & UReg
            & 33.34 & 8.29 & 48.04
            & 54.87 & 23.59 & \underline{17.86}
            & 18.76 & 16.66 & 19.08
            & \underline{48.10} & 12.82
            & 27.40 \\
            \midrule

            \multirow{7}{*}{Std.-F1}
            & \textbf{TS-Router}
            & \textbf{42.44} & \textbf{36.56} & \textbf{52.01}
            & \textbf{75.05} & \underline{29.98} & \textbf{58.71}
            & 18.83 & \textbf{40.41} & \textbf{43.88}
            & \textbf{56.41} & \textbf{70.78}
            & \textbf{47.73} \\
            & SATzilla
            & 29.63 & 8.22 & 42.78
            & 43.26 & 28.22 & 14.82
            & \textbf{22.52} & 38.81 & 22.32
            & 40.57 & 13.22
            & 27.67 \\
            & ARGOSMART
            & 28.33 & \underline{34.73} & 35.74
            & \underline{52.80} & 16.95 & \underline{17.87}
            & 20.56 & 29.39 & 23.91
            & 43.48 & \underline{21.15}
            & 29.54 \\
            & MetaOD
            & 29.75 & 8.22 & 42.54
            & 43.26 & 28.22 & 14.82
            & \textbf{22.52} & \underline{39.00} & 22.74
            & 42.44 & 12.79
            & 27.85 \\
            & ISAC
            & \underline{33.46} & \underline{34.73} & 42.38
            & 43.13 & \textbf{36.65} & 14.82
            & \underline{20.88} & 36.48 & \underline{27.80}
            & 42.07 & 13.37
            & \underline{31.43} \\
            & MSAD
            & 28.36 & 8.22 & 41.94
            & 41.75 & 16.95 & \underline{17.87}
            & 19.81 & 33.08 & 17.55
            & 45.32 & 12.43
            & 25.75 \\
            & UReg
            & 28.52 & 8.22 & \underline{42.97}
            & 43.13 & 23.46 & \underline{17.87}
            & 20.65 & 19.58 & 15.54
            & \underline{45.55} & 10.73
            & 25.11 \\
            \bottomrule
        \end{tabular}%
    }
\end{table*}
	
\end{document}